\documentclass[lettersize,journal]{IEEEtran}
\usepackage{amsmath,amsfonts}
\usepackage{algorithmic}
\usepackage{algorithm}
\usepackage{array}
\usepackage{textcomp}
\usepackage{stfloats}
\usepackage{url}
\usepackage{verbatim}
\usepackage{graphicx}
\usepackage{cite}
\usepackage{amsmath}
\usepackage{booktabs} 
\usepackage{makecell}
\usepackage[table]{xcolor}
\usepackage{multirow}
\usepackage{subfigure} 
\definecolor{Ocean}{RGB}{77,133,189}
\definecolor{Orange}{RGB}{247,144,61} 
\begin{document}

\title{SAFE: Self-Adjustment Federated Learning Framework for Remote Sensing Collaborative Perception}
\author{Xiaohe Li, Haohua Wu, Jiahao Li, Zide Fan, ~\IEEEmembership{Member,~IEEE}, Kaixin Zhang, Xinming Li, ~\IEEEmembership{Member,~IEEE}, Yunping Ge, Xinyu Zhao 
\thanks{Xiaohe Li, Haohua Wu, Jiahao Li, Zide Fan, Kaixin Zhang, Xinming Li, Yunping Ge and Xinyu Zhao are with Aerospace Information Research Institute, Chinese Academy of Sciences (e-mail: lixiaohe@aircas.ac.cn, wuhaohua23@mails.ucas.ac.cn, lijiahao24@mails.ucas.ac.cn, fanzd@aircas.ac.cn, zhangkaixin@mails.ucas.ac.cn, lixm004499@aircas.ac.cn, geyp@aircas.ac.cn, zhaoxinyu@aircas.ac.cn).}
\thanks{\textit{(corresponding author: Zide Fan)} }
\thanks{\textit{(Xiaohe Li and Haohua Wu contribute equally to the article)}}
\thanks{This work has been submitted to the IEEE for possible publication. Copyright may be transferred without notice, after which this version may no longer be accessible.}
}



\maketitle

\begin{abstract}
The rapid increase in remote sensing satellites has led to the emergence of distributed space-based observation systems. However, existing distributed remote sensing models often rely on centralized training, resulting in data leakage, communication overhead, and reduced accuracy due to data distribution discrepancies across platforms. To address these challenges, we propose the \textit{Self-Adjustment FEderated Learning} (SAFE) framework, which innovatively leverages federated learning to enhance collaborative sensing in remote sensing scenarios. SAFE introduces four key strategies: (1) \textit{Class Rectification Optimization}, which autonomously addresses class imbalance under unknown local and global distributions. (2) \textit{Feature Alignment Update}, which mitigates Non-IID data issues via locally controlled EMA updates. (3) \textit{Dual-Factor Modulation Rheostat}, which dynamically balances optimization effects during training. (4) \textit{Adaptive Context Enhancement}, which is designed to improve model performance by dynamically refining foreground regions, ensuring computational efficiency with accuracy improvement across distributed satellites. Experiments on real-world image classification and object segmentation datasets validate the effectiveness and reliability of the SAFE framework in complex remote sensing scenarios.

\end{abstract}


\section{Introduction}
In recent years, distributed space-based remote sensing observation platforms have witnessed remarkable advancements, facilitating flexible and efficient full-coverage perception of surface targets\cite{wang2023dcm,zhang2023dcnnet}. High-resolution remote sensing imagery, acquired through many distributed nodes, has been extensively employed in critical applications, including target monitoring, disaster assessment, and vegetation analysis\cite{zhang2024ffca,zhang2023superyolo,ma2024multilevel}. However, conventional distributed remote sensing methodologies predominantly rely on rigid, tightly coupled multi-terminal architectures and centralized optimization strategies. Such approaches inevitably incur substantial data transmission costs, escalating communication overhead and introducing significant risks to data privacy.

\begin{figure}[!t]
	\centering
	\includegraphics[width=3in]{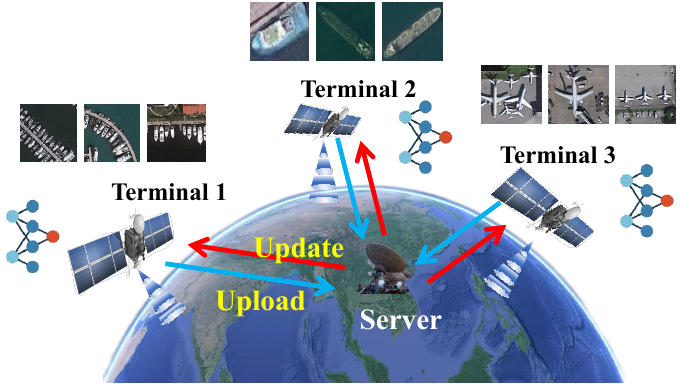}
	\caption{In the federated learning framework, each satellite terminal independently trains its model using locally collected datasets. During each communication cycle, the locally trained parameters are uploaded to a central server, which aggregates them to update the global model. The updated global parameters are then sent back to the terminals for further training.}
	\label{fig_1}
\end{figure}

Federated learning (FL), an innovative distributed training paradigm, leverages the computational capabilities of edge nodes to facilitate scalable and dynamic model training\cite{li2020review,zhang2021survey,li2020federated}. Its integration with remote sensing satellite platforms offers several significant advantages. First, edge satellite terminals can efficiently acquire localized samples, substantially enriching the diversity and volume of training datasets for diverse tasks. Second, decentralizing the training process to edge nodes eliminates the necessity of transmitting large-scale image data or feature maps, resulting in considerable reductions in communication overhead while ensuring robust privacy preservation. Furthermore, collaborative utilization of edge nodes for training optimizes resource allocation, as each satellite actively participates in training and inference processes. This approach not only alleviates the computational burden on ground-based central nodes but also expedites the deployment of operational capabilities.

In summary, developing a federated learning-based framework for collaborative remote sensing target perception demonstrates substantial application potential. As illustrated in Fig.\ref{fig_1}, each training iteration involves deploying local models on individual edge satellite nodes for training, while a central node orchestrates the global training process and manages parameter exchange. Through iterative model aggregation, this methodology enhances the performance of both local and global models, ensuring efficient and scalable learning across the distributed platform.


Despite its numerous advantages, implementing the federated learning (FL) paradigm in remote sensing introduces additional complexities compared to traditional centralized model training approaches. A primary challenge stems from the heterogeneity of training samples collected by individual satellite nodes, which exhibit substantial variations in both data sources and class distributions. Such heterogeneity often results in severe class imbalance, thereby introducing biases that can significantly degrade the performance of the global model\cite{zhuang2022divergence,makhija2022federated}.

Specifically, the imaging regions and orbital trajectories of satellite terminals vary considerably, leading to unpredictable and dynamically shifting class proportions. For instance, within the "airplane" category, the data may encompass diverse subcategories, such as large commercial aircraft (e.g., Boeing 747) and small private jets (e.g., Gulfstream series), with notable intra-class variations. These disparities create a dual imbalance challenge, as both local and global models are simultaneously affected, further complicating the optimization process. In addition, the inherently dynamic and task-specific nature of remote sensing observations exacerbates the problem, rendering conventional class imbalance mitigation strategies, such as data oversampling, weighted loss functions, or analogous techniques, largely ineffective. Typically designed for well-defined static datasets, these methods fail to address the unpredictable and evolving characteristics inherent to satellite-based remote sensing platforms.

 
Furthermore, the challenge of distributed non-independent and identically distributed (Non-IID) samples compounds the aforementioned issues\cite{li2022mocosfl}. In real-world scenarios, target categories exhibit high diversity, and due to data limitations, the gradient directions of local models demonstrate significant divergence. Even with identical initialization parameters and model architectures, substantial discrepancies in the gradient directions of local models can emerge (Fig. \ref{traj1} and Fig. \ref{traj2} depict the gradient direction differences across multiple training rounds for different clients). Existing methods attempt to mitigate this issue by transmitting sample features to facilitate unified analysis of training samples \cite{wang2024dcp} or by imposing loss function constraints to ensure the extraction of generalized features \cite{kim2023navigating}. However, these approaches often entail high communication overhead and computational costs, rendering them impractical for satellite devices with large image sizes and constrained hardware capabilities. Consequently, achieving a balance between personalization and global information during training to address the Non-IID problem remains a critical challenge.

In addition, most remote sensing images are characterized by a vertical perspective, wherein regions of interest (ROIs) containing critical contextual semantics occupy a relatively small proportion of the image. This leads to challenges such as foreground-background imbalance and scale variability. Particularly for edge satellites with limited computational resources, the careful design of network architectures on the edge side is essential to fully leverage minority foreground features, thereby enhancing the performance of remote sensing tasks such as image classification and instance segmentation.
 
Based on the aforementioned considerations, this paper proposes a novel framework, termed the \textbf{S}elf-\textbf{A}djustment \textbf{FE}derated Learning framework for Remote Sensing Collaborative Perception (SAFE), which, to the best of our knowledge, represents the first federated learning framework specifically designed to achieve both low-cost training and robust data security in remote sensing applications.

To effectively mitigate the impact of class imbalance on multi-satellite training, it is essential to monitor the training progress of different classes. Given the constrained computational and communication capabilities of edge devices, as well as privacy considerations, direct transmission of feature vectors or class information should be avoided. For instance, methods such as Sampling\cite{pouyanfar2018dynamic} or SMOTE\cite{chawla2002smote} become impractical under these constraints. Drawing inspiration from Ratio Loss\cite{wang2021addressing} and related work, we propose a low-cost \textbf{Class Rectification Optimization} (CRO) mechanism for the global model on the server side, which inherently leverages the characteristics of federated learning's training process. Specifically, during the upload phase from distributed terminals to the central server, we employ Self-Examination samples pre-stored at the central node to conduct gradient measurement for each class. By analyzing the influence of different classes on the classifier's gradients, we can assess the training progress of each class.

However, excessive class correction may lead to the loss of unique feature information in local models. As demonstrated by \cite{zhuang2022divergence} and \cite{zhuang2021collaborative}, preserving a certain degree of Non-IID information enhances the representational performance of federated learning, particularly for remote sensing satellites, where captured data often exhibit inherent uncertainty. To address this, we propose a \textbf{Feature Alignment Update} (FAU) mechanism to compare the network similarity between the server (cloud) and terminal (client) models. When the parameter differences between networks are significant, this mechanism appropriately retains the information from each client. These two mechanisms, which exhibit some inherent contradictions, are modulated by our proposed \textbf{Dual-Factor Modulation Rheostat} (DMR). Inspired by curriculum learning, this strategy prioritizes the dominant original characteristics of data samples during the initial training stages, gradually incorporating class correction effects while retaining Non-IID information as training progresses.

Additionally, to tackle challenges such as inconsistent scales across clients and foreground-background imbalance, we design an \textbf{Adaptive Context Enhancement} (ACE) module based on a lightweight multi-scale backbone. This module enhances foreground representation by applying pixel-level adjustments to identified foreground areas across clients while suppressing background regions, all executed end-to-end. This approach balances the semantic distribution of foreground and background features. It minimizes additional parameter overhead for edge-side models, enabling on-orbit satellites to perform training and inference tasks efficiently.

Overall, this work makes several significant contributions:  
\begin{itemize}
	\item This study introduces the SAFE framework, a pioneering self-adjustment federated learning approach for enhancing model training on distributed satellite networks. 
	\item The Class Rectification Optimization (CRO) strategy is introduced during the local training phase to comprehensively address the challenges of unpredictable class imbalance. During the update phase, the Feature Alignment Update (FAU) mechanism is employed to regulate the retention of local model knowledge, effectively addressing Non-IID issues. Additionally, a Dual-Factor Modulation Rheostat (DMR) strategy is designed to dynamically balance the effects of these two components.  
	\item We propose a multi-scale Adaptive Context Enhancement (ACE) method, which dynamically selects the notable foreground, maintaining computational efficiency for distributed on-orbit working while improving the overall model performance.  
	\item Experimental results on four different real-world datasets demonstrate significant improvements in image classification and object segmentation tasks, validating the SAFE framework's ability to ensure secure and reliable training in complex remote sensing scenarios.  
\end{itemize}



\section{RELATED WORK}
\subsection{Distributed Training in Remote Sensing}
With the escalating computational demands of remote sensing tasks, researchers have focused on developing distributed training methods that enable multi-platform collaboration for unified model training. Current distributed learning architectures primarily encompass centralized and decentralized paradigms, among which Federated Learning (FL) has gained significant traction in remote sensing due to its privacy-preserving mechanisms. The foundational FedAvg algorithm\cite{mcmahan2017communication} employs a parameter-averaging strategy but often encounters convergence instability in remote sensing scenarios. Subsequent advancements have improved distributed learning performance through algorithmic innovations: FedProx\cite{li2020federated} introduces a proximal term to constrain client update directions, demonstrating effectiveness in satellite image classification tasks. FedDyn\cite{jin2023feddyn} innovatively designs dynamic regularization term that adaptively adjusts constraint strength based on historical gradient information, showing superior convergence stability compared to FedProx in image classification tasks. FedNova\cite{wang2020tackling} employs a normalized weighted averaging strategy, effectively mitigating optimization biases caused by varying local iteration counts among clients. Notably, addressing the limited computational resources on satellites, Lin Hu et al. proposed a distributed collaborative learning method for few-shot remote sensing classification (FS-DCL)\cite{hu2023fs}, which achieves knowledge complementarity among satellite nodes through personalized parameter aggregation strategy and feature enhancement method, significantly enhancing feature discriminability. DCM\cite{wang2023dcm} optimizes the satellite-cloud collaborative training process, employing model fusion strategies to improve global accuracy. Ting Zhang et al. proposed a Distributed Convolutional Neural Network (DCNNet) that integrates progressive inference mechanisms with distributed self-distillation paradigms, combining multi-scale feature fusion modules and sampling-enhanced attention mechanisms, achieving breakthroughs in both efficiency and accuracy in remote sensing classification tasks\cite{zhang2023dcnnet}. However, existing methods generally assume similar class distributions across client data, which significantly contradicts the fragmented data characteristics caused by geographical distribution differences in real remote sensing scenarios. This data heterogeneity-induced model drift has become a critical bottleneck hindering the deep application of distributed learning in remote sensing.

\subsection{Non-IID Challenges in Federated Learning}
The model drift problem induced by non-independent and identically distributed (Non-IID) data heterogeneity in federated learning has spurred various innovative solutions. FedProx mitigates client deviation by constraining the optimization objective of the distance between local and global model parameters, yet its static regularization weight mechanism struggles to adapt to the dynamically changing regional data distribution characteristics in remote sensing scenarios. While Mixup\cite{han2022g} technology enhances data diversity through locally synthesized mixed samples at clients, artificially generated images may disrupt the inherent physical-spatial correlations of remote sensing features, leading to distorted feature representations. In contrast, FedU\cite{zhuang2021collaborative} builds upon the BYOL\cite{grill2020bootstrap} self-supervised learning framework to establish a dynamic interaction mechanism, effectively improving model adaptability to Non-IID data through bidirectional updates of global and local model parameters. Addressing client device heterogeneity, Hetero-SSFL\cite{makhija2022federated} innovatively allows clients to train differentiated self-supervised model architectures while aligning low-dimensional feature representation spaces using public datasets, enhancing client flexibility while providing theoretical convergence guarantees. To further optimize global aggregation efficiency, FedEMA\cite{zhuang2022divergence} introduces a dynamic exponential moving average update strategy based on model divergence, precisely capturing dynamic differences between clients and the global model by adaptively adjusting EMA decay rates, significantly enhancing model robustness in Non-IID scenarios. Meanwhile, MocoSFL\cite{li2022mocosfl} integrates Split Federated Learning with the Momentum Contrast (MoCo)\cite{he2020momentum} framework, partitioning the model into client-specific and server-shared modules, substantially reducing client computational memory overhead while enhancing knowledge transfer efficiency for Non-IID data through high-frequency synchronization and feature contrast mechanisms. The method's Target-Aware ResSFL\cite{li2022ressfl} module implements a parameter reorganization strategy that dynamically perceives target domain features, effectively resisting privacy inference attacks while reducing communication costs. Through multi-dimensional technological innovations, these methods significantly optimize resource utilization efficiency while maintaining high model accuracy, providing a practical and secure technical pathway for large-scale distributed remote sensing data processing.

\subsection{Class Imbalance Challenges}
The dual challenges of class imbalance and data heterogeneity (Non-IID) in federated learning frameworks present significant limitations in collaborative optimization approaches. Focal Loss\cite{lin2017focal} addresses class imbalance through a difficulty-aware weighting mechanism that adjusts sample loss weights, enhancing model attention to tail classes. However, its design overlooks the interference of conflicting gradient directions among clients on global aggregation in federated settings. Fed-Focal\cite{sarkar2020fed} attempts to balance data distribution by oversampling minority classes locally at clients, yet excessive reliance on repeated samples may induce local overfitting, compromising model generalization. FedLC\cite{lee2023fedlc} proposes a class-aware gradient correction strategy to harmonize parameter update directions, but its stability in complex scenarios coupling Non-IID and class imbalance remains inadequately validated. Ratio Loss\cite{wang2021addressing} innovatively integrates inter-class sample quantity ratios into loss function weight design, significantly improving tail class recall under the independent and identically distributed assumption by quantifying class scarcity. However, its core limitation lies in the heavy reliance on global class statistics for weight calculation, which, in the Non-IID environment of federated learning, often leads to overcompensation of local minority classes due to significant deviations between client-local and global class distributions, potentially exacerbating class prediction bias. Current research predominantly addresses Non-IID and class imbalance in isolation, failing to explore the intrinsic relationship between geographical distribution differences and long-tail effects in remote sensing scenarios. Considering their synergistic mechanisms, the lack of a joint optimization framework remains a critical bottleneck that hinders performance improvement in practical remote sensing applications of federated learning.

\section{METHOD}
\begin{figure*}[htbp]
	\centering
	\includegraphics[width=1\linewidth]{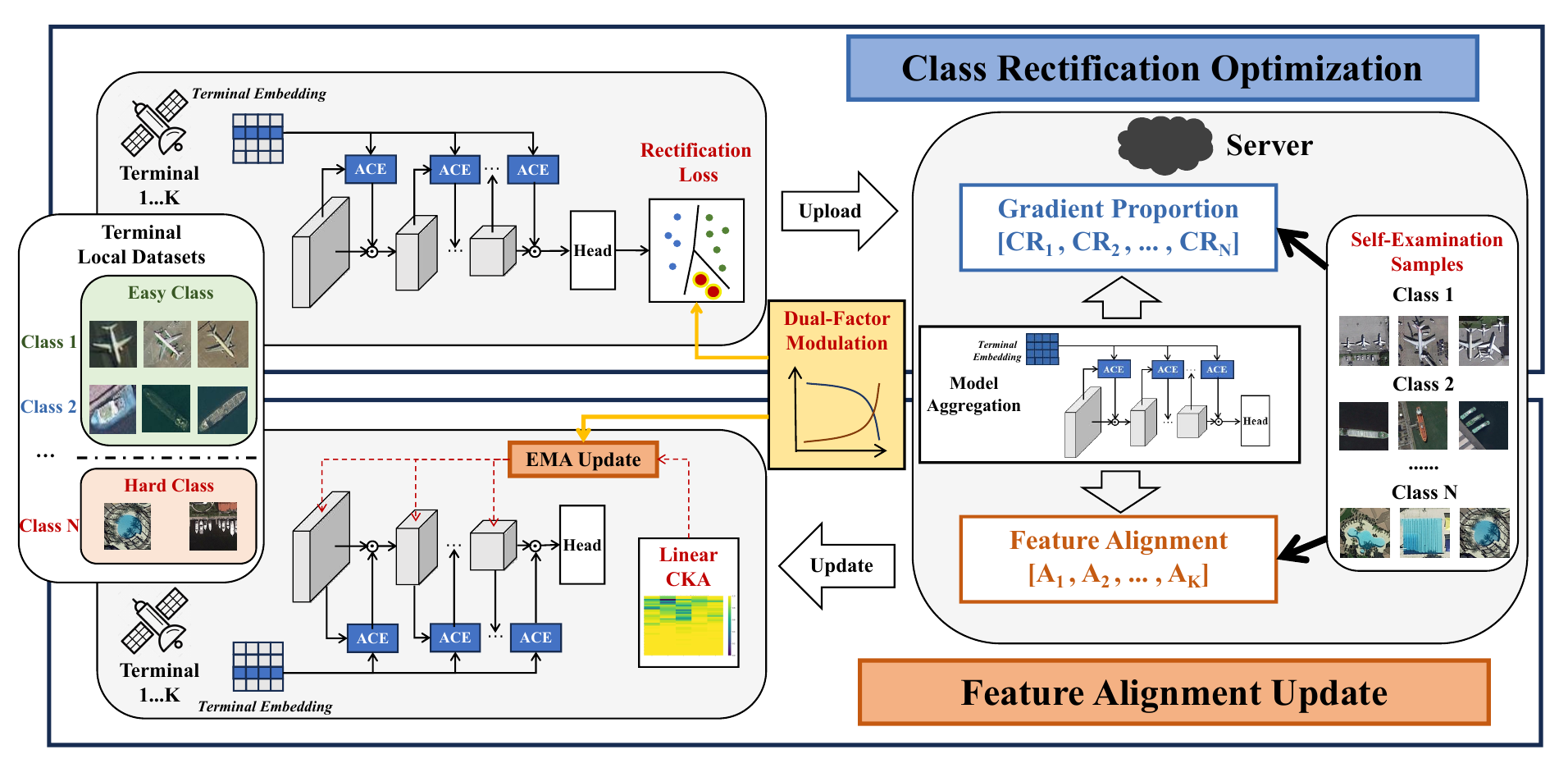}
	\caption{The proposed SAFE framework adheres to the workflow of the federated learning paradigm. A small set of Self-Examination Samples is utilized at the central node to compute Gradient Proportion and Feature Alignment. The former addresses class bias correction during local training on terminals, while the latter guides the extent of EMA updates for local models. A Dual-Factor Modulation mechanism regulates the entire training process. Additionally, the client embeddings stored on each terminal are used to guide the Adaptive Context Enhancement mechanism for foreground enhancement.}
	\label{framework}
\end{figure*}

\subsection{Overview}
The proposed SAFE framework significantly enhances the collaborative training efficiency among distributed remote sensing satellite platforms, as illustrated in Fig. \ref{framework}. Consider a satellite cluster consisting of \( K \) terminals (clients), denoted as \( \{ S_1, S_2, \dots, S_K \} \), with a total dataset \( D = \{ X, Y \} \). Within the SAFE framework, each satellite terminal \( S_i \) exclusively accesses its own local dataset \( D_i = \{ X_i^c, Y_i^c \}_{i=1}^K \). The network architecture of the \( i \)-th terminal is parameterized by \( W_i = (W_i^b, W_i^h) \), where \( W_i^b \) denotes the backbone parameters and \( W_i^h \) represents the task-specific head parameters (e.g., for image classification or instance segmentation). The prediction process is mathematically formulated as follows: $\hat{Y}_i = \mathbf{F}(W_i^b, W_i^h, X_i)$.

Our SAFE framework adheres to the standard federated learning methodology, which comprises three primary phases: (1) \textit{Local Training} on clients, (2) \textit{Model Aggregation} on the server, and (3) \textit{Model Communication} (including upload and update) between the server and clients. Initially, multiple rounds of independent model parameter optimization are conducted on each satellite. Following this, a subset of clients is randomly selected to upload their model parameters \( W_{i} \), as illustrated in the upper half of Fig. \ref{framework}. The ground-based central node subsequently performs model aggregation to derive the global model \( W_{g} \). In the second phase, the parameters of the global model are distributed to each client, allowing the integration of global information into the client models (bottom half of Fig. \ref{framework}). This strategy effectively harnesses the computational resources of edge nodes while simultaneously mitigating high communication overhead and minimizing the risk of data privacy breaches.

\subsection{Class Rectification Optimization}
We first focus on the training process of local models at each terminal. Due to variations in the activation timing and orbital dynamics of onboard cameras, the local datasets collected by individual satellites often exhibit class imbalance. Furthermore, the unique class distributions of data from each satellite lead to discrepancies between the intra-satellite class imbalance and the global data distribution.

For an image classification task with $J$ total classes, the sample count of class $j$ available on satellite terminal $S_i$ is denoted as $N_i^j$. The class distribution on terminal $S_i$ is represented as $\text{Dis}_i = [N_i^1, \dots, N_i^j, \dots, N_i^J]$, and the degree of imbalance is quantified by the ratio of the maximum to the minimum sample count, denoted as $\max(\text{Dis}_i) : \min(\text{Dis}_i)$. Similarly, the global class distribution across all terminals is defined as $\text{Dis}_g = \left[ \sum_{i=1}^K N_i^1, \dots, \sum_{i=1}^K N_i^j, \dots, \sum_{i=1}^K N_i^J \right]$, with the global imbalance expressed as $\max(\text{Dis}_g) : \min(\text{Dis}_g)$. As the adage \textit{``blind men and an elephant''} suggests, the distributions on individual terminals can significantly deviate from the global distribution, potentially leading to instability during model training after aggregation.

In training paradigms characterized by frequent parameter exchanges and model updates, adjusting class distributions solely on local models often proves ineffective in addressing global class imbalance. Consequently, corresponding adjustments should be implemented on the server (cloud) side. Conventional class adjustment methods typically rely on prior knowledge of data distributions, which may introduce risks such as communication leakage. To mitigate this issue, we propose the Class Rectification Optimization (CRO) strategy, which enables the server to independently infer class proportions without transmitting raw data or meta-information. This approach leverages gradient magnitudes to indirectly capture global class information, thereby ensuring the security and privacy of remote sensing data. Inspired by \cite{wang2021addressing}, we first establish a formal proof of the relationship between class proportions and their impact on model performance.


\textbf{Lemma 1:} For a neural network model \(F\) equipped with a nonlinear activation classifier, let the parameters of the final linear layer in the backbone model used for feature representation be denoted as \(\hat{W}\), and its output be defined as \(Z = \hat{W}X + b\). The loss function is the cross-entropy loss \(\mathcal{L}(x) = -\log(\Phi)\). If the input sample belonging to class \(p\) is denoted as \(X_p\), and the output probability of the classifier's \(q\)-th output for this sample is \(\Phi^{p}_{q}\), then let \(\bar{p}\) represent the majority class (easy samples) and \(\underline{p}\) represent the minority class (hard samples). For any third-party class \(r\) (which does not belong to \(\bar{p}\) or \(\underline{p}\)), the following inequality holds:
\[
\Phi^{\bar{p}}_{r} \leq \Phi^{\underline{p}}_{r}.
\]
Based on this result, we further derive the relationship between class proportions and gradients.

\textbf{Theorem 1:} For the aforementioned neural network model \(F\) with a nonlinear activation classifier, if the input sample of class \(p\) is denoted as \(X_p\), the gradient generated on the parameter \(\hat{W}^{p}_{q}\) of the final linear layer corresponding to class \(q\) can be expressed as:
\[
\nabla \hat{W}^{p}_{q} = \frac{\partial \mathcal{L}(X_p)}{\partial \hat{W}^{p}_{q}}.
\]
For any third-party class \(r\) (which does not belong to the majority class \(\bar{p}\) or the minority class \(\underline{p}\)), the following inequality holds:
\[
\frac{\partial \hat{W}^{\bar{p}}_{q}}{\partial \hat{W}^{\bar{p}}_{r}} \leq \frac{\partial \hat{W}^{\underline{p}}_{q}}{\partial \hat{W}^{\underline{p}}_{r}}.
\]

%

\textbf{Proof:} Assume that the output of the final linear layer is \([Z_1, \dots, Z_i, \dots, Z_C]\), where \(Z_i = \hat{W}_i \hat{X} + b\), and \(\hat{X}\) represents the forward-propagated features. Let \(\Phi_q = \text{softmax}(Z_q)\) denote the probability of the \(q\)-th class. For an input sample of class \(p\), the gradient with respect to \(\hat{W}^{p}_{q}\) (where \(p \neq q\)) can be derived as follows:

\begin{equation}
	\begin{aligned}
		\nabla \hat{W}^{p}_{q} & = \frac{\partial \mathcal{L}(X_p)}{\partial \Phi_q} \cdot \frac{\partial \Phi_q}{\partial Z_p} \cdot \frac{\partial Z_p}{\partial \hat{W}_p} \\
		& = \left(-\frac{1}{\Phi_q}\right) \cdot \left(\frac{\partial}{\partial Z_p} \frac{e^{Z_q}}{\sum_{i=1}^{C} e^{Z_i}}\right) \cdot \hat{X} \\
		& = -\left(\frac{1}{\Phi_q}\right) \cdot (\Phi_q \Phi_p) \cdot \hat{X} \\
		& = -\Phi_p \cdot \hat{X}.
	\end{aligned}
\end{equation}

In Eq. \ref{eq2.2}, the first term on the right-hand side of the first line corresponds to the derivative of the loss function, the second term represents the gradient of the softmax classifier, and the third term denotes the input \(\hat{X}\) to the final linear layer.

From the result, it can be observed that, under the same input signal \(\hat{X}\), the gradient magnitude for different classes is proportional to their corresponding probability values. Based on Lemma 1, we can approximate the following inequality:

\begin{equation}
	\label{eq2.2}
	\Phi_{\bar{p}} \cdot \hat{X}_r \geq \Phi_{\underline{p}} \cdot \hat{X}_r \Rightarrow \hat{W}^{r}_{\bar{p}} \geq \hat{W}^{r}_{\underline{p}}.
\end{equation}

According to the definitions of hard samples and easy samples, after the training process stabilizes, we have \(\Phi^{\bar{p}}_{\bar{p}} \geq \Phi^{\underline{p}}_{\underline{p}}\). It is evident that the gradient magnitude for the majority class (easy samples) is smaller. Therefore, the following inequality holds:

\begin{equation}
	\hat{W}^{r}_{\bar{p}} \cdot \hat{W}^{\underline{p}}_{\underline{p}} \geq \hat{W}^{r}_{\underline{p}} \cdot \hat{W}^{\bar{p}}_{\bar{p}}.
\end{equation}

Inspired by the aforementioned lemma, we propose to pre-store a small subset of remote sensing data from different categories on the central server. This enables the dynamic estimation of the training status for each category based on gradient information, which will be elaborated in Section ~\ref{sec:ses}. Similar to focal loss\cite{lin2017focal}, which adjusts the loss weights of samples based on their difficulty, we define the Class Rectification Loss ($\mathcal{L}_{CR}$) for class \(p\) as follows:

\begin{equation}
\label{eq4}
	CR_p = \frac{\nabla \hat{W}^{p}_{p}} {\sum^{C}_{i=1,i\neq p} \nabla \hat{W}^{i}_{p}}.
\end{equation}

\begin{equation}
\label{eq5}
	\tilde{CR} = \text{Min-Max Norm}(\{CR_i\}_{i=1}^{J}).
\end{equation}

\begin{equation}
	\mathcal{L}_{CR} = -\sum_{i=1}^{J}(\beta\cdot\tilde{CR_i} + 1)\log(\Phi_i).
\end{equation}

Here, \(CR_p\) quantifies the proportion of the gradient associated with class \(p\) to the global gradient. To ensure consistent rectification across all classes, we apply Min-Max Normalization. \(\tilde{CR}\) represents the normalized gradient proportion value for all classes. In \(\mathcal{L}_{CR}\), \(\Phi_i\) denotes the predicted probability for class \(i\), and \(\beta\) is a hyperparameter that controls the baseline adjustment for the rectification term. This loss function adjusts the coefficients of samples from different classes in the loss function based on the predicted class proportions.


\subsection{Feature Alignment Update} 
During the parameter update phase, the weights of the local models are overwritten by the global model from the central server, resulting in the loss of local information and introducing training bias in Non-IID environments. For instance, geostationary satellites capturing data from different locations may exhibit distinct local data characteristics, where insufficient local training combined with centralized information exchange can simultaneously disrupt the optimization directions of both global and local models \cite{makhija2022federated}.

Inspired by previous works\cite{zhuang2022divergence}, we propose an Exponential Moving Average (EMA)-based update strategy from terminals to the central server, termed Feature Alignment Update (FAU), to achieve self-regulated model updates. This approach facilitates the preservation of reasonable local representation capabilities.

To begin with, we focus on quantifying the discrepancy between model parameters. Direct comparison is often infeasible due to the high dimensionality of parameters. A more practical approach is to indirectly capture their similarity in the high-dimensional feature space of intermediate layers. This method is generally sensitive to network parameters while being robust to variations caused by different parameter initializations. By introducing kernel functions, we can effectively circumvent the complexity of high-dimensional mapping and computation, enabling direct calculation of distances in the high-dimensional space.

In this study, we adopt the classical Centered Kernel Alignment (CKA) method to compute the similarity of backbone networks within corresponding feature spaces at different scales. As demonstrated in \cite{kornblith2019similarity}, the representational effectiveness of network similarity under different kernel functions has been validated, and it has been shown that the Linear Kernel outperforms the RBF Kernel and Polynomial Kernel in terms of both computational efficiency and effectiveness.

In the Feature Alignment Update (FAU), we utilize a kernel function \( K = AA^T \) to incorporate the nonlinear activation layer outputs from both the server and client at various scales (denoted as \( A \in \mathbb{R}^{W \times H \times d} \)), enabling the calculation of feature extraction network similarity at corresponding scales. This can be expressed as:

\begin{equation}
\label{eq7}
	D_{CKA} = \text{CKA}(K_g, K_i) = \frac{\| A_i^T A_g \|^{2}_{F}}{\| A_g^T A_g \|_{F} \| A_i^T A_i \|_{F}},
\end{equation}

where \( D_{CKA} \) represents the divergence between the terminal and central server models, and \( \| \cdot \|_{F} \) denotes the Frobenius norm. Here, \( A_g \) corresponds to the activation layer features obtained from the backbone of the global model, while \( A_i \) represents the activation layer features extracted by the backbone of the \( i \)-th terminal model.

We propose a dynamic Exponential Moving Average (EMA)-based strategy to update the backbone parameters \( W^{b}_{i} \) of the local model as follows:

\begin{equation}
	W^{b}_{i} = \frac{1}{2}(1 - D_{CKA}) W^{b}_{i} + \frac{1}{2}(1 + D_{CKA}) W^{b}_g.
\end{equation}

The above strategy achieves similarity-based self-regulated parameter updates. The value of \( D_{CKA} \) lies in the range \([0, 1]\). When \( D_{CKA} = 1 \), the local model directly adopts the central server model \( W^{b}_g \), which aligns with the original Federated Averaging mechanism. Conversely, when \( D_{CKA} = 0 \), the local model heavily relies on its own training without incorporating external information.

\subsection{Self-Examination and Dual-Factor Modulation Rheostat}
\label{sec:ses}
The two category-related autonomous strategies are implemented through a small cloud-stored monitoring set \( D_s \), referred to as Self-Examination Samples (SES). SES comprises a small, equal number of randomly selected samples from each category, which collectively and effectively support the aforementioned Class Rectification Optimization (CRO) and Feature Alignment Update (FAU) functionalities.

During the local training phase of the model, SES belonging to \( J \) different categories are utilized to compute the gradient magnitudes, thereby facilitating the computation of \( \mathcal{L}_{CR} \).
In the update phase, in addition to transmitting parameters, feature network matrices at various scales, denoted as \( A \), are also transmitted to the clients. These matrices are utilized to compute the model's Centered Kernel Alignment (CKA) values at each client. During each communication round, the proposed autonomous strategy is applied before the model upload step to autonomously address class imbalance and is further employed in the update phase to regulate the preservation of local visual capabilities.

Inspired by curriculum learning and training strategies that progress from easy to hard samples, we propose a Dual-Factor Modulation Rheostat (DMR). It emphasizes that, during the initial stages of training, the rheostat should prioritize globally unified training. As training progresses, the framework gradually shifts its focus toward challenging class samples.

During the local training phase, the initial process emphasizes unified training across all classes. In practice, Class Rectification should play a diminished role in the early stages of training, ensuring that the model performs well across the majority of classes. As training advances, easy samples generally achieve satisfactory performance, and the loss function's weight gradually shifts toward underrepresented class samples. This ensures the model's stability and enhances the generalization capability for minority class samples on satellite terminals, even when the class distribution at the client level diverges from the global distribution. Consequently, the loss function can be formulated as:
\begin{equation}
\label{eq9}
	\mathcal{\hat{L}}_{CR} = -\sum_{i=1}^{J}(\epsilon^+\cdot\beta\cdot\tilde{CR_i} + 1)\log(\Phi_i).
\end{equation}
The function $\epsilon^+$ acts as an increasing factor that adapts with the number of training epochs, reflecting the progressively enhanced utilization of hard samples.
During the model update phase, the initial training process should fully replicate the cloud-based model to the local environment, leveraging global information to accelerate the optimization of the global model. As training progresses and the global model reaches a relatively optimal state, the proportion of the local model's weight is incrementally increased. This adjustment promotes the retention of personalized model information. The revised formula for the FAU after this adjustment is as follows:
\begin{equation}
\label{eq10}
    \begin{aligned}
        W^{b}_{i}&=\frac{1}{2}(1-\epsilon^{-}-(1-\epsilon^{-})D_{CKA})W^{b}_{i}\\
    &+\frac{1}{2}(1+\epsilon^{-}+(1-\epsilon^{-})D_{CKA})W^{b}_g.
    \end{aligned}
\end{equation}
The function $\epsilon^-$ acts as a decreasing factor that adapts with the number of training iterations, reflecting a progressively increasing utilization of hard samples.

Fig. \ref{framework} illustrates the variation in the proportions of the two loss terms during the training process. Specifically, $\epsilon^+$ can be defined as a monotonically increasing function:
\begin{equation}
\label{eq11}
\epsilon^+ = 1 - \cos\left(\frac{l}{L} \cdot \frac{\pi}{2}\right),
\end{equation}

while $\epsilon^-$ can be defined as a monotonically decreasing function:
\begin{equation}
\label{eq12}
\epsilon^- = \cos\left(\frac{l}{L} \cdot \frac{\pi}{2}\right).
\end{equation}
Both functions share the same sign for their first- and second-order derivatives. This symmetric dual-factor regulation mechanism facilitates the aforementioned process.

\begin{figure}[h]
    \centering
    \includegraphics[width=0.5\linewidth]{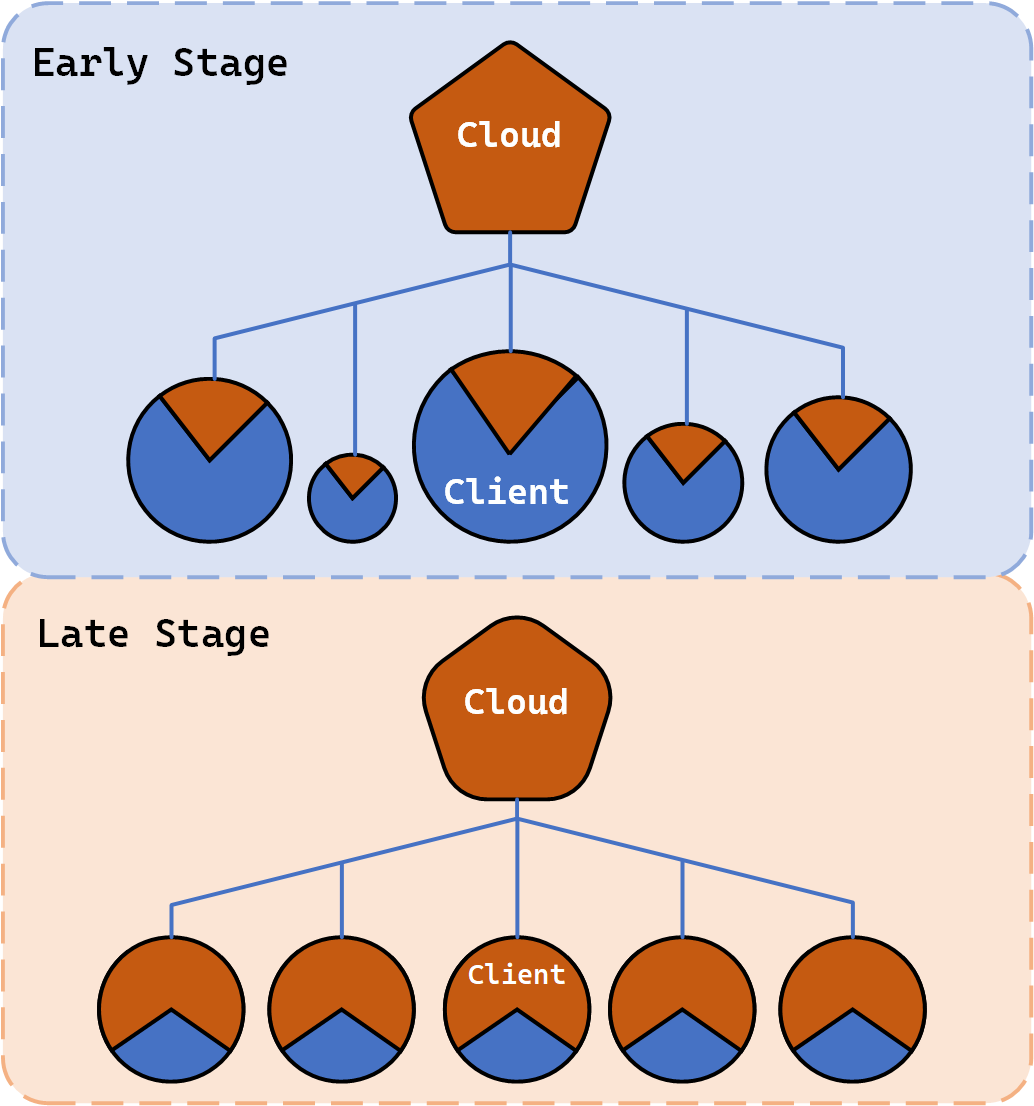}
    \caption{The Exponential Moving Average (EMA) update ratios of model parameters between clients and the cloud under the Dual-Factor Modulation Rheostat (DMR) mechanism.}
    \label{fig:dynamic_fusion}
\end{figure}

Fig.\ref{fig:dynamic_fusion} illustrates the dynamic training process of the FAU mechanism. During the initial training phase, the parameters of the client model are predominantly preserved to maximize the retention of local information, which often includes information from diverse minority classes unique to local datasets. In the later stages of training, the server model parameters become the primary focus of gradient optimization, facilitating rapid convergence to globally optimal results.

\subsection{Adaptive Context Enhancement}
\begin{figure*}[htbp]
	\centering
	\includegraphics[width=0.75\linewidth]{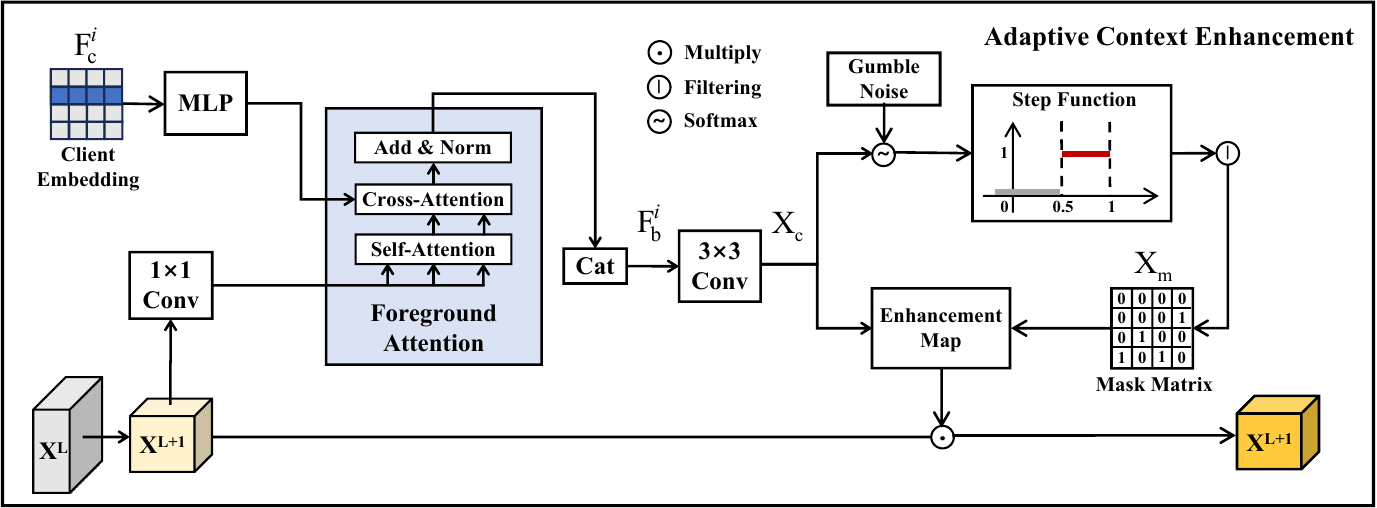}
	\caption{The proposed Adaptive Context Enhancement (ACE) module is designed to enhance the foreground regions within the feature extraction network in an end-to-end manner. Each terminal optimizes a dedicated client embedding, while cross-attention and differentiable gumbel sampling are employed to identify foreground regions and enhance them during feature forward propagation.}
	\label{ace}
\end{figure*}
Remote sensing imagery often presents challenges such as a relatively small proportion of foreground regions and inconsistent scales. However, foreground regions typically play a critical role in downstream tasks, as they often encompass complex contextual information, which directly leads to significant intra-class sample variance. For feature extraction networks (e.g., ResNet18, VGG, etc.), as previously mentioned, although multi-scale network structures are commonly employed, they tend to overlook the differences in data characteristics across different satellites. Not all patches significantly impact downstream task performance. For instance, in datasets focusing on small target observations, the foreground regions of interest (ROIs) differ substantially from those in datasets for land classification tasks. Effectively enhancing the critical foreground regions on different clients (e.g., the aircraft regions in airport imagery) can significantly improve model performance.

Du et al. propose a method that highlights foreground regions using masks in conjunction with sparse convolution \cite{du2023adaptive}. However, this approach is only applicable when the foreground bounding boxes are fully pre-defined. To achieve generalization across multi-satellite platforms, we propose an end-to-end approach that introduces trainable terminal context vectors, enabling each client to autonomously capture its foreground information while providing a global prompt capability for foreground enhancement.


As illustrated in Fig. \ref{ace}, we introduce plug-in branches at each scale of the backbone network, applying a point-wise convolution to the feature map to obtain the processed features of the original input, denoted as \( F_b^{i} \in \mathbb{R}^{W \times H \times D} \). Simultaneously, each terminal \( i \) maintains a trainable context enhancement embedding, \( F_{c}^{i} \), which represents a vectorized embedding for foreground recognition specific to that client.

During each training iteration, the central server aggregates all received context enhancement embeddings into a complete matrix \( F_{c} \in \mathbb{R}^{K \times D} \), where the \( i \)-th row corresponds to the embedding from terminal \( i \). This design is inspired by the concept of feature disentanglement. The matrix \( F_{c} \) is then passed through an MLP module for feature space transformation, followed by cross-attention with \( F_b^{i} \). The model subsequently concatenates the outputs \( F_b^{i} \) obtained from each terminal, where \( F_b^{i} \in \mathbb{R}^{W \times H \times (K \times D)} \). Following this, the concatenated features are processed through a \( 3 \times 3 \) Convolution-GN-ReLU layer to refine the foreground features, resulting in \( X_c \).

The output \( X_c \) is further converted into a differentiable 0-1 mask matrix, \( X_{m} \in \mathbb{R}^{W \times H \times 1} \), using the Gumbel-Softmax trick [35], formulated as follows:
\begin{equation}
	\begin{aligned}
	X_m' &= \phi\left(\frac{X_c + \epsilon}{\tau}\right).
	\end{aligned}
\end{equation}

\begin{equation}\label{eq2}
	X_m[i,j]=\left\{
	\begin{aligned}
		1 & , & X_m'[i,j] \geq \frac{1}{2},\\
		0 & , & X_m'[i,j] < \frac{1}{2}.
	\end{aligned}
	\right.
\end{equation}

Here, $\epsilon \in \mathbb{R}$ represents the Gumbel noise, enabling $\{0,1\}$ sampling in a differentiable manner. The enhanced recognition matrix $X_m$, composed of outputs from various terminals, demonstrates sufficient adaptability to both foreground and background regions. During training, we gradually reduce the temperature coefficient $\tau$ to make the distribution approach a one-hot representation. During inference, the mask matrix is obtained in a more relaxed manner by applying the condition $X_c \geq 0$.

For the $X_m$ matrix, a value of 0 indicates that the corresponding region is a non-critical area, and its feature intensity should be suppressed, while a value of 1 retains the intensity. The designed self-gated formula can be expressed as follows:

\begin{equation}
	X_{L} = \left(1 + X_m \cdot \frac{1}{1 + \exp(-X_m')}\right) X_{L}'.
\end{equation}
Here, \( X_{L}' \) represents the vectorized embedding obtained after feature extraction at the corresponding scale. Nonlinear activation is applied to \( X_{L}' \) to enhance the foreground, thereby endowing it with stronger representational capabilities.

It is important to note that during training, each terminal optimizes only its own context embedding, while the embeddings of other terminals are detached from the gradient optimization process. The choice of concatenation over aggregation methods, such as averaging, is inspired by the feature disentanglement mechanism. Each feature subspace focuses on its corresponding region of interest (ROI) in an interpretable manner. This design emphasizes the stable acquisition of foreground perception capabilities across all satellite platforms.

\subsection{Training Process}
The overall framework of SAFE is outlined as Algorithm~\ref{alg1} follows, incorporating three self-adjustment mechanisms that collectively address three key challenges in distributed satellite federated learning:
\begin{algorithm}[htbp]
	\caption{Training Process of SAFE}
	\label{alg1}
	\renewcommand{\algorithmicrequire}{\textbf{Input:}}
	\renewcommand{\algorithmicensure}{\textbf{Output:}}
	\begin{algorithmic}[1]
		\REQUIRE Number of clients $K$, global communication rounds $T$, local training epochs $E$, total classes $J$, with $n$ and $n_i$ representing the total training samples and samples at client $i$ respectively.
		\ENSURE Optimized server model $W_g$ and client models $W_i$.
		
		\STATE \underline{Server Execution:}
		\STATE Initialize server model parameters $W_g^b$ and $W_g^h$, maintaining Self-Examination Samples.
		\STATE Initialize class ratio estimation $\hat{CR} \leftarrow [\frac{1}{K} \text{ for } i \in [1,K]]$.
		\STATE Initialize model divergence $D_{CKA} \leftarrow 1$.
		
		\FOR{each communication round $t \in [1,T]$}
			\STATE Randomly select $k$ clients $S_k$ from $K$ available clients.
			\FOR{each client $k \in S_k$}
				\STATE $W_i^b, W^h_i \leftarrow \mathrm{ClientUpdate}(W^b, W^h, t, \hat{CR}, D_{CKA})$.
				\STATE Update gradient proportion $\hat{CR}$ using Eq. \ref{eq4} and Eq. \ref{eq5}.
				\STATE Update model divergence $D_{CKA}$ using Eq. \ref{eq7}.
			\ENDFOR
			\STATE Aggregate model parameters:
			\STATE $W_g^b \leftarrow \sum_{k \in S_k}\frac{n_k}{n}W_i^b$.
			\STATE $W_g^h \leftarrow \sum_{k \in S_k}\frac{n_k}{n}W_i^h$.
		\ENDFOR
		
		\STATE \underline{ClientUpdate}$(W^b, W^h, t, \hat{CR}, D_{CKA})$:
		\FOR{each local epoch $e \in [1,E]$}
			\STATE Compute $\epsilon^{+}$ and $\epsilon^{-}$ using Eq. \ref{eq11} and Eq. \ref{eq12}.
			\STATE Update model parameters $W$ using Eq. \ref{eq10}.
			\STATE Optimize $W_i^b$ and $W^h_i$ using Eq. \ref{eq9}.
		\ENDFOR
	\end{algorithmic}
\end{algorithm}

\section{EXPERIMENTS}
In the experimental section, we aim to address three fundamental challenges to thoroughly evaluate the efficacy and adaptability of the proposed Self-Adjustment Federated Learning (SAFE) framework. The experimental design is structured to prove the following advantages:

\begin{enumerate}
    \item \textbf{Advancement Comparison:} SAFE's superiority is validated via state-of-the-art comparisons, emphasizing accuracy, efficiency, stability, and robustness to Non-IID data and class imbalance in remote sensing.  

    \item \textbf{Module Mechanisms:} Ablation studies dissect SAFE's core strategies (e.g., Class Rectification, Feature Alignment), revealing their roles in resolving data heterogeneity, class imbalance, and sample scarcity across distributed platforms.  

    \item \textbf{Adaptability Analysis:} Key parameter and data setting impacts are systematically evaluated, demonstrating SAFE's adaptability and robustness in diverse remote sensing scenarios.
\end{enumerate}

Through these three experimental components, we aim to thoroughly demonstrate the innovation, effectiveness, and superior performance of the SAFE framework in complex remote sensing scenarios, offering new insights and a solid foundation for future collaborative sensing and distributed model design in remote sensing applications.

\subsection{Experimental Setup}
This section is organized into five subsections. Initially, the experimental dataset is introduced. Subsequently, the experimental configuration and environment are elaborated. Following this, the evaluation methodology employed in this study is delineated. Moreover, the baseline models utilized for comparative analysis are specified. Ultimately, pertinent experimental particulars are emphasized.

\subsubsection{Dataset}
To thoroughly assess the performance of our proposed method in remote sensing image classification and segmentation tasks, extensive experiments are conducted on two classification datasets, namely PatternNet\cite{zhou2018patternnet} and NWPU-RESISC45\cite{cheng2017remote}, as well as two semantic segmentation datasets, including LoveDA\cite{wang2021loveda} and WHDLD\cite{shao2018performance}.

\textbf{PatternNet:} This dataset is another extensively utilized benchmark for remote sensing image classification, consisting of 30,400 images spanning 38 categories, with 800 images per category. The images are sourced from Google Earth and Google Maps, primarily covering US cities. Each image has a fixed size of $256\times256$ pixels, with spatial resolutions varying between 0.062 and 4.693 meters per pixel. PatternNet incorporates fine-grained categories, such as "basketball court" and "tennis court," thereby increasing the complexity of the classification task.

\textbf{NWPU-RESISC45:} This dataset constitutes a large-scale benchmark for remote sensing image classification, comprising 31,500 images distributed across 45 categories, collected from over 100 countries and regions. Each category contains 700 images with a resolution of $256\times256$ pixels, while the spatial resolution ranges from approximately 30 to 0.2 meters per pixel for most scene classes. Compared to existing datasets, NWPU-RESISC45 demonstrates greater intra-class diversity and inter-class similarity. For instance, categories such as "sparse residential" and "dense residential" exhibit semantic overlap, presenting significant challenges for accurate classification.

\textbf{LoveDA:} This dataset is designed explicitly for domain-adaptive semantic segmentation in high-resolution remote sensing imagery. It comprises 5,987 high-resolution images ($1024 \times 1024$ pixels) with 166,768 annotated objects collected from three distinct cities in China. Compared to existing datasets, LoveDA incorporates two different domains (urban and rural), introducing challenges such as multi-scale objects, complex background samples, and imbalanced class distributions.

\textbf{WHDLD:} WHDLD (WuHan Dense Labeled Dataset) is designed for dense semantic segmentation in urban remote sensing scenarios. It comprises 4,940 high-resolution RGB images captured by Gaofen-1 and Ziyuan-3 satellite sensors over Wuhan, China. Each image is standardized to a size of 256 × 256 pixels with a spatial resolution of 2 meters, enabling fine-grained analysis of urban land cover. The annotations are densely labelled across six key categories: bare land, building, sidewalk, water body, vegetation, and road.

\begin{figure*}[htbp]
\centering
\subfigure{
\begin{minipage}[t]{0.25\linewidth}
\centering
\includegraphics[width=1.8in]{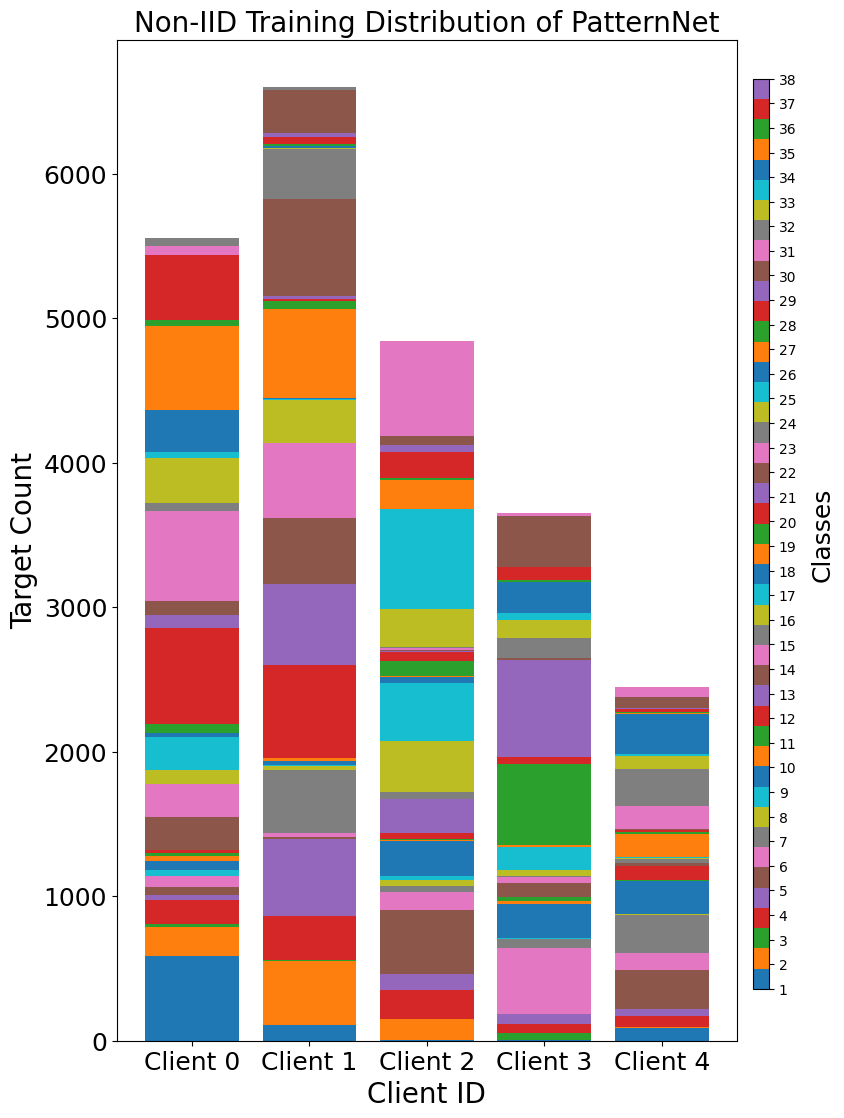}
\end{minipage}%
}%
\subfigure{
\begin{minipage}[t]{0.25\linewidth}
\centering
\includegraphics[width=1.8in]{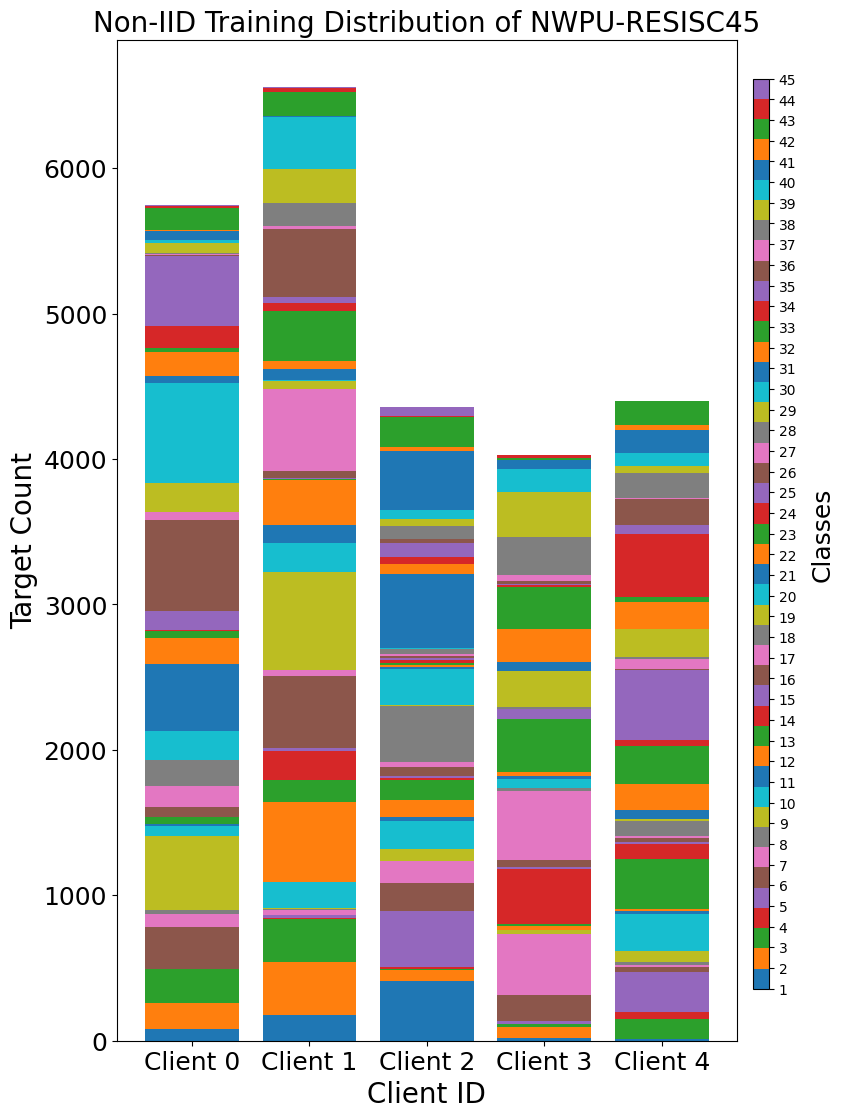}
\end{minipage}%
}%
\subfigure{
\begin{minipage}[t]{0.25\linewidth}
\centering
\includegraphics[width=1.8in]{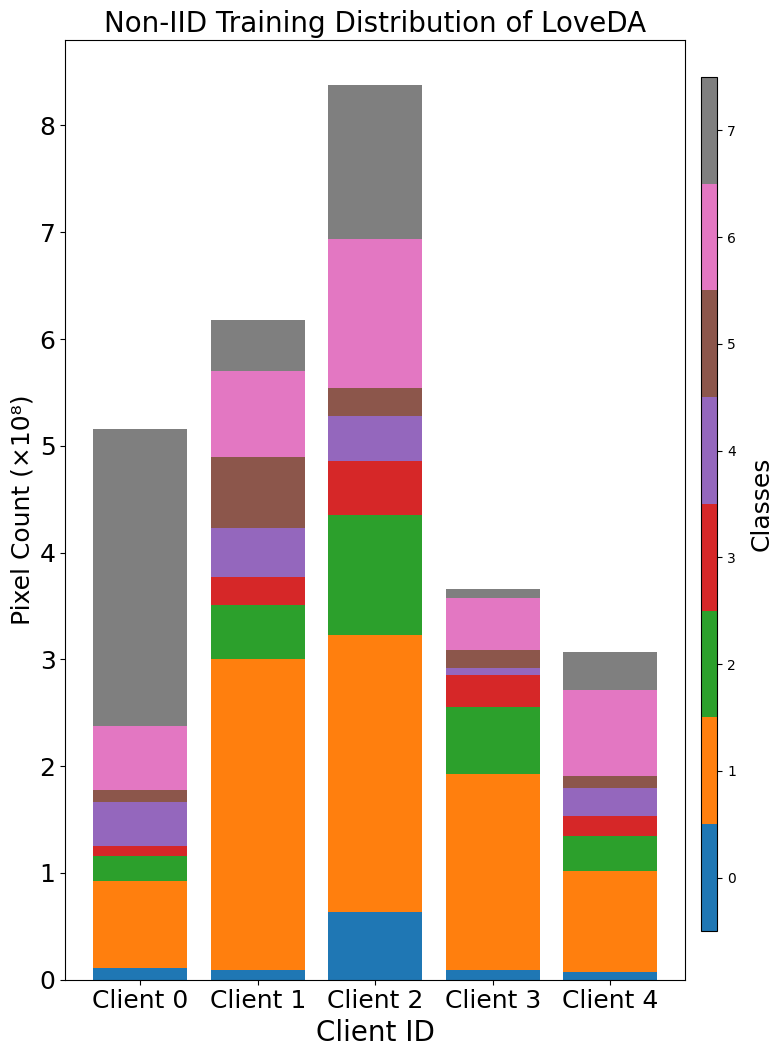}
\end{minipage}%
}%
\subfigure{
\begin{minipage}[t]{0.25\linewidth}
\centering
\includegraphics[width=1.8in]{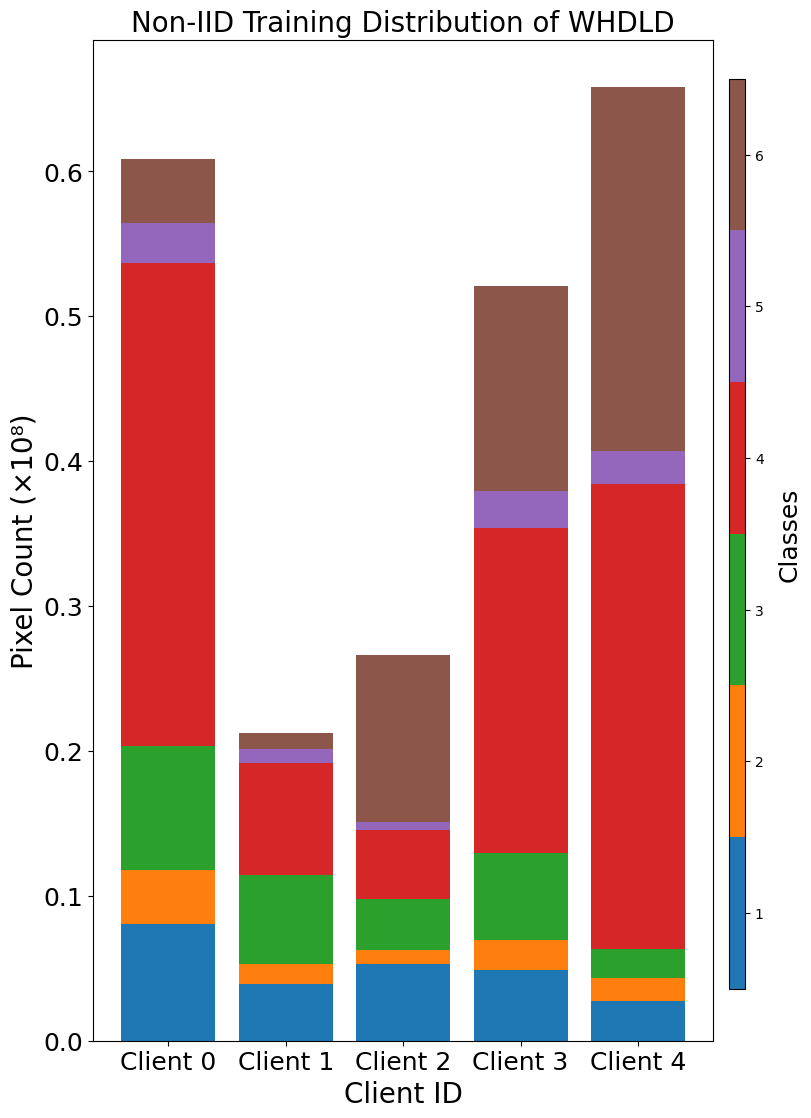}
\end{minipage}%
}%
\centering
\caption{The distribution of classes across clients in the PatternNet, NWPU-RESISC45, LoveDA and WHDLD datasets.}
\label{dataset}
\end{figure*}

Data collected by different terminals often exhibits imbalanced class distributions in real-world scenarios, deviating from uniform or independent and identically distributed (IID) patterns. To simulate practical Non-IID data distributions, we implement a dual-phase preprocessing pipeline for distributed data partitioning of NWPU-RESISC45 and PatternNet: 1) strategic class-wise sample reduction to induce natural imbalance, followed by 2) Dirichlet distribution-driven client allocation that preserves real-world data imbalance patterns. Segmentation datasets are inherently imbalanced at the pixel level. Therefore, we argue that random data distribution across clients is sufficient to simulate real-world environmental conditions. The class distribution across different terminals for the four datasets is illustrated in Fig. \ref{dataset}.


\subsubsection{Implementation Details}
In our experiments, the initial learning rate is set to $1 \times 10^{-4}$ and decays progressively during training. The model is trained for 200 epochs with a batch size of 64. We adopt the Adam optimizer for optimization and employ the cross-entropy loss function for gradient computation. The dataset is divided into training and validation sets with an 80\%-20\% split ratio, where 80\% of the data is used for training, and the remaining 20\% is reserved for validation. 
We assume five clients participate in the training process by default, with an imbalanced class ratio of 10:1. The $\beta$ coefficient in the class loss function is set to 1.0.
All experiments are implemented using PyTorch 1.12, CUDA 11.6, and an A800 GPU with 80GB memory.

\subsubsection{Evaluation Metrics}
For classification tasks, we employ two primary metrics: sample accuracy and class accuracy. Sample accuracy measures global prediction performance as the ratio of correctly identified instances to total instances, ranging from 0 to 1. While effective for balanced datasets, it may overestimate performance in imbalanced scenarios due to the dominance of majority classes. Class accuracy evaluates per-category performance by computing the proportion of correctly classified samples within each class. The mean class accuracy averaged across all classes mitigates majority class bias, providing a robust measure for imbalanced datasets. Higher per-class accuracy indicates improved discriminative power for the corresponding category. For segmentation tasks, we utilize the mean intersection over union (mIoU), a widely adopted metric that quantifies the overlap between predicted and ground truth targets. The IoU metric is defined as the overlap area divided by the union area of prediction and annotation, bounded within [0, 1]. mIoU extends this by averaging IoU values across all target categories, delivering a holistic assessment of segmentation consistency.

\subsection{Comparison with Existing Methods}
This section conducts comparative experiments to evaluate the effectiveness and robustness of the proposed SAFE framework against three categories of baselines:

\begin{enumerate}
    \item In the federated learning context, SAFE is compared with FedAvg\cite{mcmahan2017communication}, FedProx\cite{li2020federated}, FedDyn\cite{jin2023feddyn}, and FedNova\cite{wang2020tackling}. FedAvg represents the most fundamental and classic framework in federated learning. FedProx mitigates client heterogeneity via proximal regularization, FedDyn addresses Non-IID data through dynamic regularization, and FedNova reduces data heterogeneity by normalizing the averaging process. This comparison assesses SAFE's robustness under heterogeneous and Non-IID data, which are common challenges in remote sensing.
    \item SAFE is benchmarked against lightweight models (MobileNet V2\cite{sandler2018mobilenetv2}, MobileNet V3s/V3l\cite{howard2019searching,nugroho2023improved}, ShuffleNetV2\cite{ma2018shufflenet} and EfficientNet\cite{tan2019efficientnet}) optimized for edge deployment. They also employ FedAvg for distributed training without any improvement applied to the network architecture. This comparison demonstrates SAFE's ability to balance computational efficiency and high performance in resource-constrained scenarios.
    \item The impact of different backbones is analyzed by replacing ResNet-18\cite{he2016deep} with VGG\cite{sengupta2019going}. This exploration highlights the trade-offs between model complexity, generalization, and performance under heterogeneous data distributions, particularly relevant for remote sensing applications.
\end{enumerate}

To comprehensively evaluate SAFE, experiments compare its performance with existing methods on both classification and segmentation tasks.

As shown in Table \ref{tab:method_comparison}, we validate the performance advantages of the SAFE framework under typical Non-IID scenarios on the classification datasets PatternNet and NWPU-RESISC45. The client data's class distribution and Non-IID scenarios are constructed using the Dirichlet distribution ($\alpha=0.5$).

In the table, we list the class accuracy (C.Acc) and sample accuracy (S.Acc) achieved by different methods on both the server and client sides. Notably, C.Acc effectively reflects the model's capability to handle long-tailed distributions. The results demonstrate that the SAFE framework, with ResNet-18 as the backbone network, achieves significant breakthroughs in cloud-class accuracy (Cloud C.Acc): compared to FedAvg, improvements of $17.92\%$ ($71.35\% \rightarrow 89.27\%$) and $16.95\%$ $(48.25\% \rightarrow 65.20\%)$ are observed on PatternNet and NWPU-RESISC45, respectively. Even when compared to other federated learning improvements such as FedDyn, the enhancements remain substantial, reaching $15.25\%$ (PatternNet) and $14.06\%$ (NWPU). These breakthroughs validate the synergistic effects of the FAU and CRO mechanisms in mitigating multi-satellite data distribution discrepancies.

Traditional federated learning methods perform poorly due to their neglect of remote sensing data characteristics. FedAvg and FedProx achieve client-class accuracy (Client C.Acc) of only $38.92\%-39.87\%$ on the NWPU dataset, significantly lagging behind SAFE ($52.76\%$). This gap stems from their static aggregation strategies, which lack adaptability to scenarios with limited sample sizes. Lightweight baseline models, such as MobileNetV3l, fall behind SAFE by $16.31\%$ in client C.Acc due to inherent network architecture limitations, while EfficientNet performs even worse. The SAFE network enhances feature representation through the ACE mechanism without significantly increasing computational or parameter costs, demonstrating strong practical applicability.

Cross-dataset performance comparisons reveal the SAFE framework's adaptive capability to the complexity of remote sensing scenarios. Compared to FedAvg, SAFE achieves a more pronounced improvement in cloud metrics on the NWPU-RESISC45 dataset than on PatternNet, attributed to the former's multi-source heterogeneous characteristics and geographical distribution diversity, which exacerbate feature distribution discrepancies among clients and pose greater challenges to traditional methods. Notably, SAFE exhibits higher client sample accuracy $(95.87\%, 81.92\%)$ than cloud metrics $(94.84\%, 76.49\%)$ on both datasets. This phenomenon arises because the client test sets follow the same distribution as their training sets, simulating the regional characteristics monitored by satellites, whereas the cloud test set aggregates data from all clients, simulating a global target distribution. Leveraging the Dual-Factor Modulation Rheostat mechanism, SAFE preserves more client-side parameters in the early training stages to reduce gradient oscillation and shifts toward preserving more cloud-side parameters in later stages. This design trade-off enables SAFE to maintain personalized perception capabilities at the satellite terminal level while enhancing generalization for global targets through cloud aggregation (class accuracy improves significantly from $69.93\%$ to $89.27\%$). This achieves a federated learning framework with "local specialization and global balance" for collaborative remote sensing.

In Table \ref{tab:method_comparison_segmentation}, we also compare the performance of SAFE with other methods on segmentation tasks using the classic U-Net\cite{ronneberger2015u}+ResNet-50 architecture on the LoveDA and WHDLD datasets. SAFE demonstrates strong adaptability to segmentation tasks, outperforming other federated learning methods by $10\%-15\%$ in mIoU metrics. Since segmentation tasks lack significant Non-IID class discrepancies and the cloud aggregates models from all clients, cloud performance is generally superior. Among other federated learning methods, FedNova and FedDyn perform slightly better.

\label{tab:method_comparison}
\begin{table*}[t]
    \centering
    \caption{Performance comparison of different methods on distributed classification tasks. All results are averaged over five runs, with the best outcomes highlighted in bold and the second-best results underlined. Color blocks indicate the ranking of different backbones for comparison.}
    \resizebox{0.9\linewidth}{!}{
    \small
    \begin{tabular}{l|c|c|c|c|c|c|c|c|c|c}
        \toprule
        Method & \multicolumn{4}{c|}{PatternNet} & \multicolumn{4}{c}{NWPU-RESISC45} & \multicolumn{2}{c}{Complexity}\\
        \midrule
        & \makecell[c]{Cloud \\ C.Acc\%} & \makecell[c]{Client \\ C.Acc\%} & \makecell[c]{Cloud \\ S.Acc\%} & \makecell[c]{Client \\ S.Acc\%} 
        & \makecell[c]{Cloud \\ C.Acc\%} & \makecell[c]{Client \\ C.Acc\%} & \makecell[c]{Cloud \\ S.Acc\%} & \makecell[c]{Client \\ S.Acc\%} & \makecell[c]{FLOPS(G)} & \makecell[c]{Params(M)}\\
        \midrule
        FedAvg(ResNet-18) & \cellcolor{Ocean!30}{71.35} & \cellcolor{Ocean!30}{56.40} & \cellcolor{Ocean!40}{87.91} & \cellcolor{Ocean!50}{91.02} 
               & \cellcolor{Ocean!20}{48.25} & \cellcolor{Ocean!10}{38.92} & \cellcolor{Ocean!30}{66.82} & \cellcolor{Ocean!30}{72.14} &13.24 & 11.19\\ 
        FedProx(ResNet-18) & \cellcolor{Ocean!40}{72.14} & \cellcolor{Ocean!40}{56.62} & \cellcolor{Ocean!50}{88.38} & \cellcolor{Ocean!60}{91.24} 
                & \cellcolor{Ocean!30}{49.30} & \cellcolor{Ocean!20}{39.87} & \cellcolor{Ocean!40}{67.91} & \cellcolor{Ocean!40}{73.45} &- & -\\
        FedDyn(ResNet-18) & \cellcolor{Ocean!60}{{74.02}} & \cellcolor{Ocean!60}{{58.12}} & \cellcolor{Ocean!70}{{89.14}} & \cellcolor{Ocean!80}{{92.57}} 
                & \cellcolor{Ocean!50}{{51.14}} & \cellcolor{Ocean!50}{{41.32}} & \cellcolor{Ocean!60}{{69.43}} & \cellcolor{Ocean!70}{{75.82}} &- & -\\
        FedNova(ResNet-18) & \cellcolor{Ocean!50}{73.76} & \cellcolor{Ocean!50}{57.89} & \cellcolor{Ocean!60}{89.02} & \cellcolor{Ocean!70}{92.41} 
                & \cellcolor{Ocean!40}{50.87} & \cellcolor{Ocean!40}{41.05} & \cellcolor{Ocean!50}{69.12} & \cellcolor{Ocean!60}{75.47} &- & -\\
        \hline
        MobilenetV2 & \cellcolor{Ocean!10}{66.99} & \cellcolor{Ocean!10}{50.96} & \cellcolor{Ocean!20}{86.81} & \cellcolor{Ocean!20}{87.56} 
                    & \cellcolor{Ocean!10}{46.80} & \cellcolor{Ocean!05}{33.45} & \cellcolor{Ocean!10}{58.43} & \cellcolor{Ocean!10}{66.48} &2.37 &2.27\\
        MobilenetV3s & \cellcolor{Ocean!20}{69.41} & \cellcolor{Ocean!20}{53.21} & \cellcolor{Ocean!30}{87.09} & \cellcolor{Ocean!30}{89.24} 
                    & \cellcolor{Ocean!15}{47.92} & \cellcolor{Ocean!10}{35.74} & \cellcolor{Ocean!20}{60.32} & \cellcolor{Ocean!20}{68.75} &1.52 &0.93\\
        MobilenetV3l & \cellcolor{Ocean!25}{70.85} & \cellcolor{Ocean!25}{54.37} & \cellcolor{Ocean!35}{87.92} & \cellcolor{Ocean!35}{89.75} 
                     & \cellcolor{Ocean!18}{48.60} & \cellcolor{Ocean!12}{36.45} & \cellcolor{Ocean!22}{61.05} & \cellcolor{Ocean!22}{69.20} &5.45 &1.65\\
        ShuffleNetV2 & \cellcolor{Ocean!15}{64.33} & \cellcolor{Ocean!08}{49.20} & \cellcolor{Ocean!15}{84.58} & \cellcolor{Ocean!15}{86.10} 
                     & \cellcolor{Ocean!08}{44.25} & \cellcolor{Ocean!04}{31.81} & \cellcolor{Ocean!08}{56.37} & \cellcolor{Ocean!08}{64.00} &1.43 &0.59\\
        EfficientNet & \cellcolor{Ocean!05}{62.45} & \cellcolor{Ocean!05}{47.83} & \cellcolor{Ocean!10}{83.12} & \cellcolor{Ocean!10}{85.74} 
                     & \cellcolor{Ocean!05}{42.67} & \cellcolor{Ocean!02}{30.23} & \cellcolor{Ocean!05}{54.76} & \cellcolor{Ocean!05}{61.89} & 5.35&3.98\\
        \midrule
        SAFE (VGG-11) & \cellcolor{Ocean!70}{\underline{79.40}} & \cellcolor{Ocean!70}{\underline{62.64}} & \cellcolor{Ocean!80}{\underline{92.58}} & \cellcolor{Ocean!90}{\underline{95.29}} 
                      & \cellcolor{Ocean!60}{\underline{59.20}} & \cellcolor{Ocean!60}{\underline{48.99}} & \cellcolor{Ocean!70}{\underline{71.29}} & \cellcolor{Ocean!90}{\textbf{82.57}} &53.65 &128.92\\
        SAFE (ResNet-18) & \cellcolor{Ocean!90}{\textbf{89.27}} & \cellcolor{Ocean!90}{\textbf{69.93}} & \cellcolor{Ocean!100}{\textbf{94.84}} & \cellcolor{Ocean!95}{\textbf{95.87}} 
                         & \cellcolor{Ocean!80}{\textbf{65.20}} & \cellcolor{Ocean!80}{\textbf{52.76}} & \cellcolor{Ocean!90}{\textbf{76.49}} & \cellcolor{Ocean!85}{\underline{81.92}} &14.76 & 13.13 \\
        \bottomrule
    \end{tabular}
    }
    \label{tab:method_comparison}
\end{table*}

\begin{table*}[t]
    \centering
    \caption{Performance comparison of different methods on distributed segmentation tasks. All results are averaged over five runs, with the best outcomes highlighted in bold and the second-best results underlined. Color blocks indicate the ranking of different backbones for comparison.}
    \resizebox{0.68\linewidth}{!}{
    \small
    \begin{tabular}{l|c|c|c|c}
        \toprule
        Method & \multicolumn{2}{c|}{LoveDA} & \multicolumn{2}{c}{WHDLD} \\
        \midrule
        & Client mIoU\% & Cloud mIoU\% & Client mIoU\% & Cloud mIoU\% \\
        \midrule
        FedAvg(ResNet-50) & \cellcolor{Ocean!25}{40.89} & \cellcolor{Ocean!10}{43.15} & \cellcolor{Ocean!10}{52.42} & \cellcolor{Ocean!10}{55.30} \\
        FedProx(ResNet-50) & \cellcolor{Ocean!10}{40.83} & \cellcolor{Ocean!25}{43.25} & \cellcolor{Ocean!25}{53.63} & \cellcolor{Ocean!25}{56.17} \\
        FedDyn(ResNet-50) & \cellcolor{Ocean!40}{41.27} & \cellcolor{Ocean!55}{44.83} & \cellcolor{Ocean!40}{56.35} & \cellcolor{Ocean!55}{59.29} \\
        FedNova(ResNet-50) & \cellcolor{Ocean!55}{41.56} & \cellcolor{Ocean!40}{44.12} & \cellcolor{Ocean!55}{56.84} & \cellcolor{Ocean!40}{58.58} \\  
        \midrule
        SAFE (VGG-11) & \cellcolor{Ocean!85}{\underline{46.77}} & \cellcolor{Ocean!70}{\underline{48.02}} & \cellcolor{Ocean!70}{\underline{58.95}} & \cellcolor{Ocean!70}{\underline{62.55}} \\
        SAFE (ResNet-50) & \cellcolor{Ocean!70}{\textbf{46.62}} & \cellcolor{Ocean!85}{\textbf{48.24}} & \cellcolor{Ocean!85}{\textbf{59.78}} & \cellcolor{Ocean!85}{\textbf{63.22}} \\
        \bottomrule
    \end{tabular}
    }
    \label{tab:method_comparison_segmentation}
\end{table*}

\subsection{Adaptability Analysis}

In this section, we evaluate the adaptability of the proposed SAFE framework by analyzing the impact of key parameters and data settings on its performance. Specifically, we focus on the following aspects:

\subsubsection{The Quantity of Clients}
The number of clients significantly impacts federated learning performance. By varying client participation, we assess its effects on convergence speed, communication overhead, and model accuracy, evaluating SAFE's efficiency under diverse network configurations. As shown in Table \ref{tab:client_scaling}, SAFE demonstrates robust adaptability to data sparsity as the number of clients (C) increases from $2$ to $20$. On PatternNet, cloud class accuracy (Cloud C.Acc) decreases by $25.95\%$ ($94.81\%\rightarrow68.86\%$), while on NWPU-RESISC45, the decrease is $19.33\%$ ($79.48\%\rightarrow60.15\%$). Notably, when $C=20$, SAFE significantly outperforms FedAvg across multiple metrics. This scalability highlights SAFE's ability to mitigate model degradation from data fragmentation through CRO and FAU mechanisms, adapting to collaborative learning needs across small ($C=2-10$) to large-scale ($C=20+$) satellite clusters.

\subsubsection{The Ratio of Class Imbalance}
We examine SAFE's performance under varying class imbalance ratios, particularly setting $10:1$ when $(C=5)$, to evaluate its robustness in handling imbalanced data. Table \ref{tab:class_imbalance} highlights SAFE's adaptability to long-tailed distributions. As the imbalance ratio increases from $2:1$ to $20:1$, Cloud C.Acc on NWPU-RESISC45 decreases by $13.99\%$ ($74.51\%\rightarrow60.52\%$), while on PatternNet, it only drops $7.36\%$ ($94.18\%\rightarrow86.82\%$). This stability is attributed to the CRO mechanism's selective enhancement of tail classes. SAFE also outperforms baseline methods, with improvements of $27.87\%$ and $28.74\%$ in Cloud C.Acc on the two datasets, respectively.

Notably, SAFE's sample accuracy remains stable as the imbalance ratio increases. On NWPU-RESISC45, client S.Acc only slightly decreases from $81.20\%$ to $80.69\%$ (fluctuation $\textless0.6\%$). This robustness stems from the coupling effect of class distribution characteristics and optimization strategies. Head classes dominate sample accuracy, while the CRO strategy dynamically compensates for tail classes through loss weighting. SAFE's sample-level robustness aligns with the common remote sensing requirement of "precise recognition of rare but critical classes."

\subsubsection{The Coefficient $\beta$ of Class Rectification Optimization (CRO)}
The CRO strategy mitigates class imbalance and enhances model performance. We analyze the impact of varying CRO coefficients $\beta$ on accuracy and convergence to identify optimal settings for remote sensing tasks. Table \ref{tab:cro_coefficients} shows that CRO coefficients must be configured based on data complexity. PatternNet peaks at a coefficient of $10$ with $89.27\%$, but performance drops by $0.66$ percentage points at a coefficient of $20$, as excessively high coefficients reduce the gradient contribution of head classes. On NWPU-RESISC45, Cloud C.Acc improves by $6.27$ percentage points at a coefficient of $20$ ($65.35\%$) compared to a coefficient of $0.4$, confirming that complex scenarios (cross-region, multi-resolution) require stronger class compensation. Notably, as the CRO coefficient increases, sample accuracy declines slightly, as the CRO mechanism selectively enhances tail classes through dynamic loss weighting, affecting other classes to some extent. This verifies the optimization mechanism of "sacrificing minor precision of majority classes to significantly improve minority classes."

\subsubsection{The Frequency of Feature Alignment Update (FAU)}
The frequency of parameter update from cloud to client significantly impacts the degree of model divergence. By adjusting training epochs between two updates, we study its effect on training efficiency and accuracy under different data distributions, optimizing the balance between local computation and global synchronization. Table \ref{tab:fau_periods} shows that as the FAU period increases, global and local consistency declines, reducing classification accuracy (C.Acc) and sample accuracy (S.Acc). On PatternNet, increasing the FAU period from $1$ to $10$ decreases cloud class accuracy (Cloud C.Acc) from $93.39\%$ to $84.73\%$ and client class accuracy (Client C.Acc) from $75.33\%$ to $68.18\%$, with drops of $8.66\%$ and $7.15\%$ percentage points, respectively. Cloud and client sample accuracy also decline, indicating delayed model updates and reduced global model adaptability to data distribution changes. This trend is consistent on NWPU-RESISC45. Increasing the FAU period decreases cloud class accuracy from $77.81\%$ to $59.00\%$ (a drop of $18.81\%$) and client class accuracy from $63.64\%$ to $46.62\%$ (a drop of $17.02\%$), highlighting the impact of reduced FAU frequency on uneven data distribution. Prolonged lack of alignment adjustments degrades classification performance. Frequent FAU updates mitigate model drift caused by uneven data distribution, improving classification accuracy. However, increased FAU frequency raises communication overhead, necessitating a trade-off between communication efficiency and accuracy. To balance this trade-off, we set FAU period = $2-5$. On NWPU-RESISC45, cloud class accuracy reaches $65.20\%$, lower than FAU period = $1$ ($77.81\%$) but significantly higher than FAU period = $10$ ($59.00\%$). Communication costs at FAU period = $5$ are reduced by $80\%$ compared to FAU period = $1$, alleviating communication pressure.

\begin{table*}[t]
    \centering
    \caption{The performance of the model under varying quantities of clients.}
    \resizebox{0.7\linewidth}{!}{
    \small
    \begin{tabular}{l|l|c|c|c|c|c|c|c|c}
        \toprule
        \multicolumn{2}{l|}{Settings} & \multicolumn{4}{c|}{PatternNet} & \multicolumn{4}{c}{NWPU-RESISC45} \\
       \cmidrule(lr){1-2} \cmidrule(lr){3-6} \cmidrule(lr){7-10}
        Method & Number & \makecell[c]{Cloud\\C.Acc\%} & \makecell[c]{Client\\C.Acc\%} & \makecell[c]{Cloud\\S.Acc\%} & \makecell[c]{Client\\S.Acc\%} 
        & \makecell[c]{Cloud\\C.Acc\%} & \makecell[c]{Client\\C.Acc\%} & \makecell[c]{Cloud\\S.Acc\%} & \makecell[c]{Client\\S.Acc\%} \\
        \midrule
        \multirow{4}{*}{SAFE} & C=2    & \cellcolor{Ocean!85}{\textbf{94.81}} & \cellcolor{Ocean!85}{\textbf{84.53}} & \cellcolor{Ocean!85}{\textbf{98.00}} & \cellcolor{Ocean!85}{\textbf{98.21}} 
               & \cellcolor{Ocean!85}{\textbf{79.48}} & \cellcolor{Ocean!85}{\textbf{75.87}} & \cellcolor{Ocean!85}{\textbf{87.59}} & \cellcolor{Ocean!85}{\textbf{88.89}} \\
        & C=5    & \cellcolor{Ocean!65}{\underline{89.27}} & \cellcolor{Ocean!65}{\underline{69.93}} & \cellcolor{Ocean!65}{\underline{94.84}} & \cellcolor{Ocean!65}{\underline{95.87}} & \cellcolor{Ocean!65}{\underline{65.20}} & \cellcolor{Ocean!65}{\underline{52.76}} & \cellcolor{Ocean!65}{\underline{76.49}} & \cellcolor{Ocean!65}{\underline{81.92}} \\
        & C=10   & \cellcolor{Ocean!45}{83.60} & \cellcolor{Ocean!45}{64.81} & \cellcolor{Ocean!45}{92.79} & \cellcolor{Ocean!45}{93.48}
               & \cellcolor{Ocean!45}{62.37} & \cellcolor{Ocean!45}{48.31} & \cellcolor{Ocean!45}{72.84} & \cellcolor{Ocean!45}{76.19} \\
        & C=20   & \cellcolor{Ocean!25}{68.86} & \cellcolor{Ocean!25}{48.84} & \cellcolor{Ocean!25}{86.73} & \cellcolor{Ocean!25}{87.46} 
               & \cellcolor{Ocean!25}{60.15} & \cellcolor{Ocean!25}{45.48} & \cellcolor{Ocean!25}{70.29} & \cellcolor{Ocean!25}{73.36} \\
        \midrule
        FedAvg& C=20 & \cellcolor{Ocean!10}{61.34} & \cellcolor{Ocean!10}{39.95}  & \cellcolor{Ocean!10}{80.46} & \cellcolor{Ocean!10}{79.14} & \cellcolor{Ocean!10}{52.30} & \cellcolor{Ocean!10}{32.90} & \cellcolor{Ocean!10}{71.26} & \cellcolor{Ocean!10}{72.15}  \\
        \bottomrule
    \end{tabular}
    }
    \label{tab:client_scaling}
\end{table*}

\begin{table*}[t]
    \centering
    \caption{The performance of the model under varying ratio of class imbalance.}
    \resizebox{0.7\linewidth}{!}{
    \small
    \begin{tabular}{l|l|c|c|c|c|c|c|c|c}
        \toprule
        \multicolumn{2}{l|}{Settings} & \multicolumn{4}{c|}{PatternNet} & \multicolumn{4}{c}{NWPU-RESISC45} \\
       \cmidrule(lr){1-2} \cmidrule(lr){3-6} \cmidrule(lr){7-10}
       Method & Number  & \makecell[c]{Cloud\\C.Acc\%} & \makecell[c]{Client\\C.Acc\%} & \makecell[c]{Cloud\\S.Acc\%} & \makecell[c]{Client\\S.Acc\%} 
        & \makecell[c]{Cloud\\C.Acc\%} & \makecell[c]{Client\\C.Acc\%} & \makecell[c]{Cloud\\S.Acc\%} & \makecell[c]{Client\\S.Acc\%} \\
        \midrule
        \multirow{4}{*}{SAFE}  & 2:1   & \cellcolor{Ocean!85}{\textbf{94.18}} & \cellcolor{Ocean!85}{\textbf{81.78}} & \cellcolor{Ocean!85}{\textbf{95.33}} & \cellcolor{Ocean!85}{\textbf{96.15}} 
              & \cellcolor{Ocean!85}{\textbf{74.51}} & \cellcolor{Ocean!85}{\textbf{58.81}} & \cellcolor{Ocean!65}{\underline{77.49}} & \cellcolor{Ocean!65}{\underline{81.20}} \\
        & 5:1   & \cellcolor{Ocean!65}{\underline{91.73}} & \cellcolor{Ocean!65}{\underline{75.86}} & \cellcolor{Ocean!65}{\underline{95.08}} & \cellcolor{Ocean!65}{\underline{96.01}} 
              & \cellcolor{Ocean!65}{\underline{66.79}} & \cellcolor{Ocean!65}{\underline{56.10}} & \cellcolor{Ocean!85}{\textbf{77.81}} & \cellcolor{Ocean!25}{80.29} \\
        & 10:1  & \cellcolor{Ocean!45}{89.27} & \cellcolor{Ocean!45}{69.93} & \cellcolor{Ocean!45}{94.84} & \cellcolor{Ocean!45}{95.87} 
              & \cellcolor{Ocean!45}{65.20} & \cellcolor{Ocean!45}{52.76} & \cellcolor{Ocean!45}{76.49} & \cellcolor{Ocean!85}{\textbf{81.92}} \\
        & 20:1  & \cellcolor{Ocean!25}{86.82} & \cellcolor{Ocean!25}{64.00} & \cellcolor{Ocean!25}{94.60} & \cellcolor{Ocean!25}{95.73} 
              & \cellcolor{Ocean!25}{60.52} & \cellcolor{Ocean!25}{50.19} & \cellcolor{Ocean!25}{70.42} & \cellcolor{Ocean!45}{80.69} \\
        \midrule
        FedAvg &  20:1        & \cellcolor{Ocean!10}{66.31}     & \cellcolor{Ocean!10}{56.09}     & \cellcolor{Ocean!10}{88.41}     & \cellcolor{Ocean!10}{90.40}     & \cellcolor{Ocean!10}{45.77}     & \cellcolor{Ocean!10}{35.43}     & \cellcolor{Ocean!10}{61.89}     & \cellcolor{Ocean!10}{67.70}     \\
        \bottomrule
    \end{tabular}
    }
    \label{tab:class_imbalance}
\end{table*}

\begin{table*}[t]
    \centering
    \caption{The performance of the model under varying coefficients $\beta$ of Class Rectification Optimization (CRO).}
    \resizebox{0.7\linewidth}{!}{
    \small
    \begin{tabular}{l|c|c|c|c|c|c|c|c}
        \toprule
         Settings& \multicolumn{4}{c|}{PatternNet} & \multicolumn{4}{c}{NWPU-RESISC45} \\
        \cmidrule(lr){1-1} \cmidrule(lr){2-5} \cmidrule(lr){6-9}
        Coefficient & \makecell[c]{Cloud\\C.Acc\%} & \makecell[c]{Client\\C.Acc\%} & \makecell[c]{Cloud\\S.Acc\%} & \makecell[c]{Client\\S.Acc\%} 
        & \makecell[c]{Cloud\\C.Acc\%} & \makecell[c]{Client\\C.Acc\%} & \makecell[c]{Cloud\\S.Acc\%} & \makecell[c]{Client\\S.Acc\%} \\
        \midrule
        $\beta$ = 0.4      & \cellcolor{Ocean!20}{88.19} & \cellcolor{Ocean!40}{70.18} & \cellcolor{Ocean!80}{\textbf{95.34}} & \cellcolor{Ocean!80}{\textbf{96.88}} 
               & \cellcolor{Ocean!20}{59.08} & \cellcolor{Ocean!20}{51.84} & \cellcolor{Ocean!80}{\textbf{78.03}} & \cellcolor{Ocean!80}{\textbf{82.89}} \\
        $\beta$ = 0.6      & \cellcolor{Ocean!60}{\underline{88.84}} & \cellcolor{Ocean!80}{\textbf{71.38}} & \cellcolor{Ocean!60}{\underline{95.23}} & \cellcolor{Ocean!60}{\underline{96.53}} 
               & \cellcolor{Ocean!40}{60.20} & \cellcolor{Ocean!60}{\underline{52.84}} & \cellcolor{Ocean!60}{\underline{77.01}} & \cellcolor{Ocean!60}{\underline{82.59}} \\
        $\beta$ = 0.8     & \cellcolor{Ocean!80}{\textbf{89.27}} & \cellcolor{Ocean!20}{69.93} & \cellcolor{Ocean!40}{94.84} & \cellcolor{Ocean!20}{95.87} 
               & \cellcolor{Ocean!60}{\underline{65.20}} & \cellcolor{Ocean!40}{52.76} & \cellcolor{Ocean!40}{76.49} & \cellcolor{Ocean!40}{81.92} \\
        $\beta$ = 1.0     & \cellcolor{Ocean!40}{88.61} & \cellcolor{Ocean!60}{\underline{71.34}} & \cellcolor{Ocean!20}{94.45} & \cellcolor{Ocean!40}{96.03} 
               & \cellcolor{Ocean!80}{\textbf{65.35}} & \cellcolor{Ocean!80}{\textbf{52.85}} & \cellcolor{Ocean!20}{74.43} & \cellcolor{Ocean!20}{80.15} \\
        \bottomrule
    \end{tabular}
    }
    \label{tab:cro_coefficients}
\end{table*}

\begin{table*}[t]
    \centering
    \caption{The performance of the model under varying frequency of Feature Alignment Update (FAU).}
    \resizebox{0.7\linewidth}{!}{
    \small
    \begin{tabular}{l|c|c|c|c|c|c|c|c}
        \toprule
        Settings&  \multicolumn{4}{c|}{PatternNet} & \multicolumn{4}{c}{NWPU-RESISC45} \\
        \cmidrule(lr){1-1}  \cmidrule(lr){2-5} \cmidrule(lr){6-9}
        Period & \makecell[c]{Cloud\\C.Acc\%} & \makecell[c]{Client\\C.Acc\%} & \makecell[c]{Cloud\\S.Acc\%} & \makecell[c]{Client\\S.Acc\%} 
        & \makecell[c]{Cloud\\C.Acc\%} & \makecell[c]{Client\\C.Acc\%} & \makecell[c]{Cloud\\S.Acc\%} & \makecell[c]{Client\\S.Acc\%} \\
        \midrule
        1      & \cellcolor{Ocean!80}{\textbf{93.39}} & \cellcolor{Ocean!80}{\textbf{75.33}} & \cellcolor{Ocean!80}{\textbf{97.50}} & \cellcolor{Ocean!80}{\textbf{97.77}} 
               & \cellcolor{Ocean!80}{\textbf{77.81}} & \cellcolor{Ocean!80}{\textbf{63.64}} & \cellcolor{Ocean!80}{\textbf{85.25}} & \cellcolor{Ocean!80}{\textbf{87.59}} \\
        2      & \cellcolor{Ocean!60}{\underline{92.74}} & \cellcolor{Ocean!60}{\underline{73.35}} & \cellcolor{Ocean!60}{\underline{96.46}} & \cellcolor{Ocean!60}{\underline{97.23}} 
               & \cellcolor{Ocean!60}{\underline{72.77}} & \cellcolor{Ocean!60}{\underline{57.42}} & \cellcolor{Ocean!60}{\underline{81.14}} & \cellcolor{Ocean!60}{\underline{84.40}} \\
        5      & \cellcolor{Ocean!40}{89.27} & \cellcolor{Ocean!40}{69.93} & \cellcolor{Ocean!40}{94.84} & \cellcolor{Ocean!40}{95.87} 
               & \cellcolor{Ocean!40}{65.20} & \cellcolor{Ocean!40}{52.76} & \cellcolor{Ocean!40}{76.49} & \cellcolor{Ocean!40}{81.92} \\
        10     & \cellcolor{Ocean!20}{84.73} & \cellcolor{Ocean!20}{68.18} & \cellcolor{Ocean!20}{92.48} & \cellcolor{Ocean!20}{95.62} 
               & \cellcolor{Ocean!20}{59.00} & \cellcolor{Ocean!20}{46.62} & \cellcolor{Ocean!20}{70.57} & \cellcolor{Ocean!20}{79.71} \\
        \bottomrule
    \end{tabular}
    }
    \label{tab:fau_periods}
\end{table*}

\begin{figure*}[t]  
	\centering
	\includegraphics[width=0.8\linewidth]{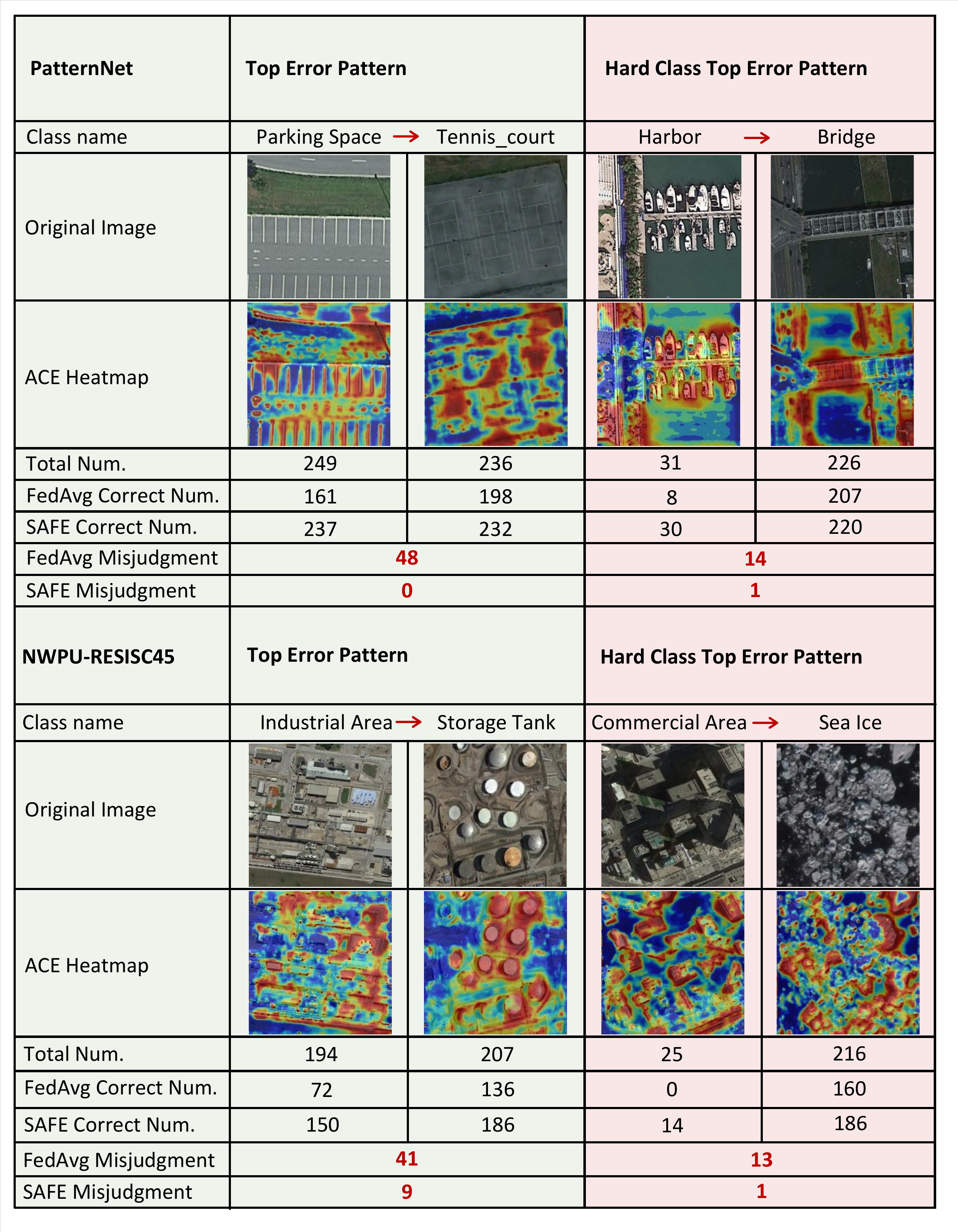}  
	\caption{Quantitative and qualitative analysis on typical misclassification cases in the PatternNet and NWPU-RESISC45 datasets.}  
	\label{fig:case_study}
\end{figure*}
\subsection{Ablation Study}

In this section, we conduct ablation experiments on the three key modules of the SAFE framework—Feature Alignment Update (FAU), Class Rectification Optimization (CRO), and Adaptive Context Enhancement (ACE)—to evaluate their contributions to model performance. Table \ref{tab:cross_dataset_ablation} and Table \ref{tab:segmentation_ablation} present the progressive integration results of these modules with the baseline ResNet model, ultimately leading to the full SAFE framework.

The FAU module significantly mitigates performance degradation caused by data heterogeneity. On PatternNet, client-class accuracy increases from $56.40\%$ to $66.42\%$, while on NWPU-RESISC45, it improves from $33.92\%$ to $50.73\%$. Corresponding cloud-class accuracy rose to $74.89\%$ and $63.21\%$, respectively, confirming FAU's effectiveness in addressing client-side data distribution discrepancies. For segmentation tasks, similar improvements are observed: comparative analysis of mIoU (mean Intersection over Union) between server and terminal models reveals that the FAU mechanism achieves notable enhancements, primarily by preserving personalized information during parameter updates.

With the incorporation of the CRO module, performance further improves. Cloud class accuracy on PatternNet increases from $74.89\%$ to $89.23\%$, and on NWPU-RESISC45 from $63.21\%$ to $65.20\%$, highlighting CRO's advantages in handling class imbalance, particularly on the more challenging NWPU-RESISC45 dataset. In segmentation applications, however, CRO's benefits are less pronounced due to the difficulty of explicit artificial augmentation of challenging classes, though modest improvements were still observed on client-side models.

The ACE module, when combined with FAU, enhances client-model accuracy. On PatternNet, client-class accuracy increases from $66.42\%$ (FAU alone) to $69.11\%$, demonstrating ACE's capability to amplify critical foreground features and improve terminal model expressiveness. For segmentation tasks, the integration of ACE enhances foreground feature representations, enabling more effective entity segmentation. Compared to CRO, ACE delivers additional improvements of $2–3\%$ in segmentation accuracy.

By synergistically integrating FAU, CRO, and ACE, the complete SAFE framework achieves $89.27\%$ cloud-class accuracy on PatternNet with stable client accuracy at $69.93\%$. On NWPU-RESISC45, the framework demonstrates robust performance, significantly surpassing both individual modules and the baseline model. For segmentation benchmarks, state-of-the-art performance is consistently achieved on LoveDA and WHDLD datasets. These results validate the necessity of coordinated module integration to address critical challenges in remote sensing tasks, including data heterogeneity and class imbalance.

\begin{table*}[t]
    \centering
    \caption{Ablation study on PatternNet and NWPU-RESISC45.}
    \resizebox{0.72\linewidth}{!}{ 
    \small
    \begin{tabular}{l|c|c|c|c|c|c|c|c}
        \toprule
        Method & \multicolumn{4}{c|}{PatternNet} & \multicolumn{4}{c}{NWPU-RESISC45} \\
        \cmidrule(lr){2-5} \cmidrule(lr){6-9}
        & \makecell[c]{Cloud\\C.Acc\%} & \makecell[c]{Client\\C.Acc\%} & \makecell[c]{Cloud\\S.Acc\%} & \makecell[c]{Client\\S.Acc\%} 
        & \makecell[c]{Cloud\\C.Acc\%} & \makecell[c]{Client\\C.Acc\%} & \makecell[c]{Cloud\\S.Acc\%} & \makecell[c]{Client\\S.Acc\%} \\
        \midrule
        ResNet-18 (base)  & \cellcolor{Ocean!10}{71.35} & \cellcolor{Ocean!10}{56.40} & \cellcolor{Ocean!10}{87.91} & \cellcolor{Ocean!10}{91.02} 
                          & \cellcolor{Ocean!10}{47.08} & \cellcolor{Ocean!10}{33.92} & \cellcolor{Ocean!10}{58.83} & \cellcolor{Ocean!10}{67.35} \\
        +FAU             & \cellcolor{Ocean!30}{74.89} & \cellcolor{Ocean!30}{66.42} & \cellcolor{Ocean!30}{90.61} & \cellcolor{Ocean!30}{95.85} 
                          & \cellcolor{Ocean!30}{61.97} & \cellcolor{Ocean!30}{50.12} & \cellcolor{Ocean!30}{76.21} & \cellcolor{Ocean!30}{81.67} \\
        +FAU + CRO       & \cellcolor{Ocean!70}{\underline{89.23}} & \cellcolor{Ocean!80}{\textbf{70.54}} & \cellcolor{Ocean!50}{94.59} & \cellcolor{Ocean!70}{\underline{96.03}} 
                          & \cellcolor{Ocean!50}{63.21} & \cellcolor{Ocean!50}{50.73} & \cellcolor{Ocean!70}{\underline{78.44}} & \cellcolor{Ocean!70}{\underline{83.19}} \\
        +FAU + ACE       & \cellcolor{Ocean!50}{85.26} & \cellcolor{Ocean!50}{69.11} & \cellcolor{Ocean!80}{\textbf{95.04}} & \cellcolor{Ocean!80}{\textbf{96.49}} 
                          & \cellcolor{Ocean!70}{\underline{64.20}} & \cellcolor{Ocean!70}{\underline{52.45}} & \cellcolor{Ocean!80}{\textbf{79.22}} & \cellcolor{Ocean!80}{\textbf{83.56}} \\
        SAFE            & \cellcolor{Ocean!80}{\textbf{89.27}} & \cellcolor{Ocean!70}{\underline{69.93}} & \cellcolor{Ocean!70}{\underline{94.84}} & \cellcolor{Ocean!50}{95.87} 
                          & \cellcolor{Ocean!80}{\textbf{65.20}} & \cellcolor{Ocean!80}{\textbf{52.76}} & \cellcolor{Ocean!50}{76.49} & \cellcolor{Ocean!50}{81.92} \\
        \bottomrule
    \end{tabular}
    }
    \label{tab:cross_dataset_ablation}
\end{table*}

\begin{table}[t]
    \centering
    \caption{Ablation study on LoveDA and WHDLD.}
    \resizebox{1\linewidth}{!}{
    \small
    \begin{tabular}{l|c|c|c|c}
        \toprule
        Method & \multicolumn{2}{c|}{LoveDA} & \multicolumn{2}{c}{WHDLD} \\
        \cmidrule(lr){2-3} \cmidrule(lr){4-5}
        & \makecell[c]{Cloud\\MIoU\%} & \makecell[c]{Client\\MIoU\%} 
        & \makecell[c]{Cloud\\MIoU\%} & \makecell[c]{Client\\MIoU\%} \\
        \midrule
        ResNet-50 (base)  & \cellcolor{Ocean!10}{43.15} & \cellcolor{Ocean!10}{40.89} & \cellcolor{Ocean!10}{55.30} & \cellcolor{Ocean!10}{52.42} \\
        +FAU              & \cellcolor{Ocean!30}{45.68} & \cellcolor{Ocean!30}{42.32} & \cellcolor{Ocean!30}{57.51} & \cellcolor{Ocean!30}{54.31} \\
        +FAU + CRO        & \cellcolor{Ocean!50}{45.90} & \cellcolor{Ocean!50}{43.42} & \cellcolor{Ocean!50}{59.87} & \cellcolor{Ocean!50}{58.98} \\
        +FAU + ACE        & \cellcolor{Ocean!70}{\underline{47.98}} & \cellcolor{Ocean!70}{\underline{46.11}} & \cellcolor{Ocean!70}{\underline{62.12}} & \cellcolor{Ocean!80}{\textbf{61.54}} \\
        SAFE              & \cellcolor{Ocean!80}{\textbf{48.24}} & \cellcolor{Ocean!80}{\textbf{46.62}} & \cellcolor{Ocean!80}{\textbf{63.22}} & \cellcolor{Ocean!70}{\underline{59.78}} \\
        \bottomrule
    \end{tabular}
    }
    \label{tab:segmentation_ablation}
\end{table}

\subsection{Case Study}
Fig.\ref{fig:case_study} presents a qualitative analysis of representative cases across different datasets, focusing on the most frequent misclassification patterns in PatternNet and NWPU-RESISC45 datasets. Specifically, we visualize cases where Parking Space is misclassified as Tennis Court and Industrial Area as Storage Tank, as well as challenging classes with the highest misclassification rates, such as Harbor being misclassified as Bridge and Commercial Area as Sea Ice. Additionally, attention heatmaps illustrate the attention values assigned to foreground regions by the ACE mechanism. In the figure, "Total Num." denotes the total number of samples per class, "Correct Num." represents the number of correctly identified samples during inference, and "Misjudgement" indicates the frequency of the listed misclassification patterns for the source class.

The proposed SAFE framework demonstrates the ability to accurately identify regions most relevant to each class. For instance, in optical imagery where parking spaces and tennis courts exhibit visual similarities, SAFE effectively focuses on distinguishing features such as lane markings in parking areas and court patterns. In the case of commercial areas and sea ice, where shadow and sea surface regions are challenging to differentiate due to imaging conditions, SAFE's attention heatmaps reveal sharper focus on building structures in commercial areas and smoother attention distributions for glacial features, aligning with their respective characteristics. This foreground enhancement significantly improves the perceptual accuracy of the model.

Quantitatively, the SAFE framework exhibits minimal to no errors for error-prone classes, even handling hard classes effectively. For example, in the Harbor class, FedAvg correctly identifies only 8 samples, while SAFE achieves near-perfect accuracy. For extremely challenging classes like Commercial Area, SAFE achieves over $50\%$ accuracy. For classes with larger sample sizes, SAFE consistently delivers high accuracy. These results robustly validate the effectiveness of our proposed mechanisms in training models across diverse clients, even in scenarios with severe class imbalance.


\subsection{Additional Analysis}

\subsubsection{Confusion Matrix}
\begin{figure*}[htbp]

\centering
\subfigure{
\begin{minipage}[t]{0.5\linewidth}
\centering
\includegraphics[width=3.7in]{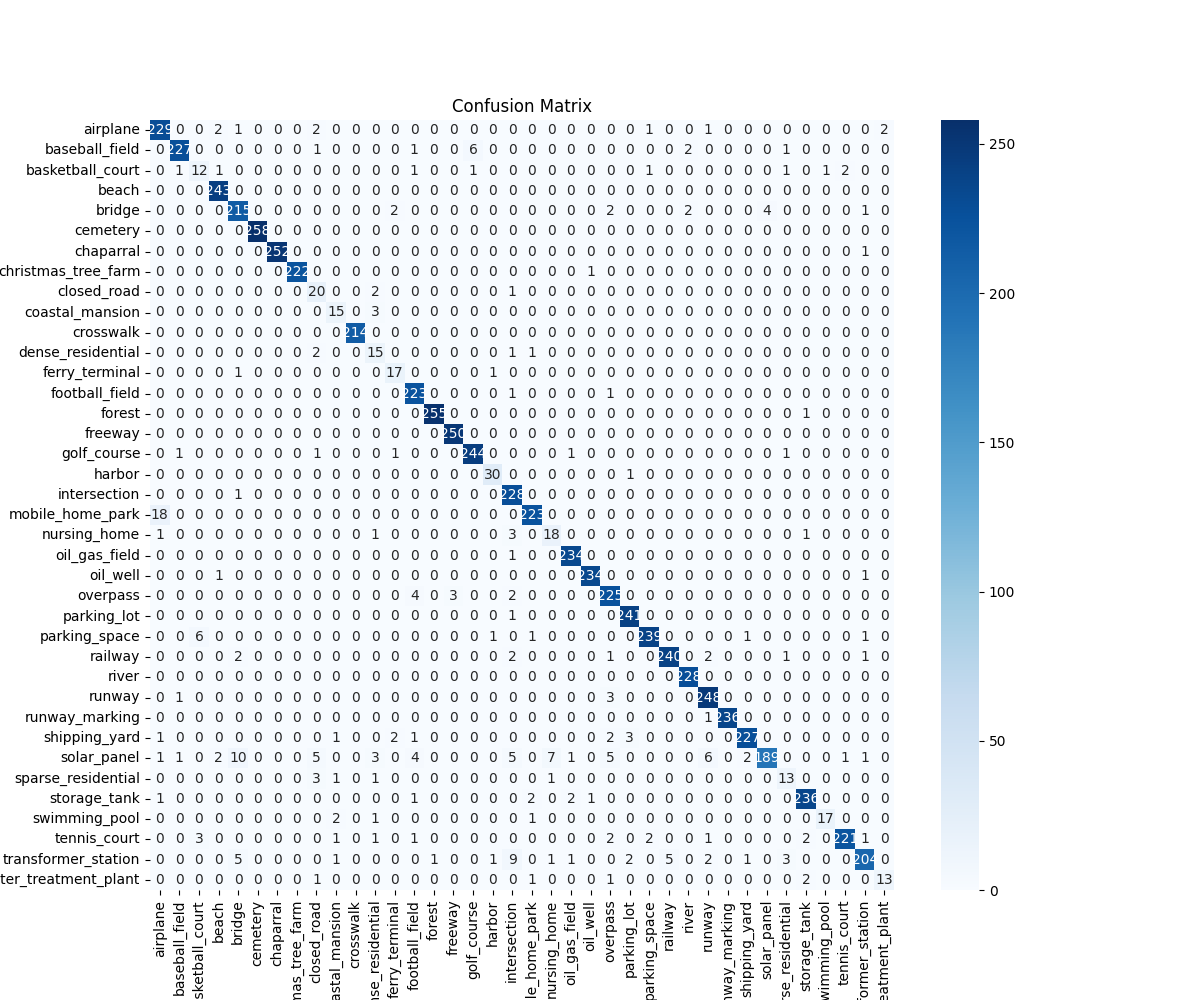}
\caption{The confusion matrix results of PatternNet under the SAFE method.}
\label{confusion1}
\end{minipage}%
}%
\subfigure{
\begin{minipage}[t]{0.5\linewidth}
\centering
\includegraphics[width=3.7in]{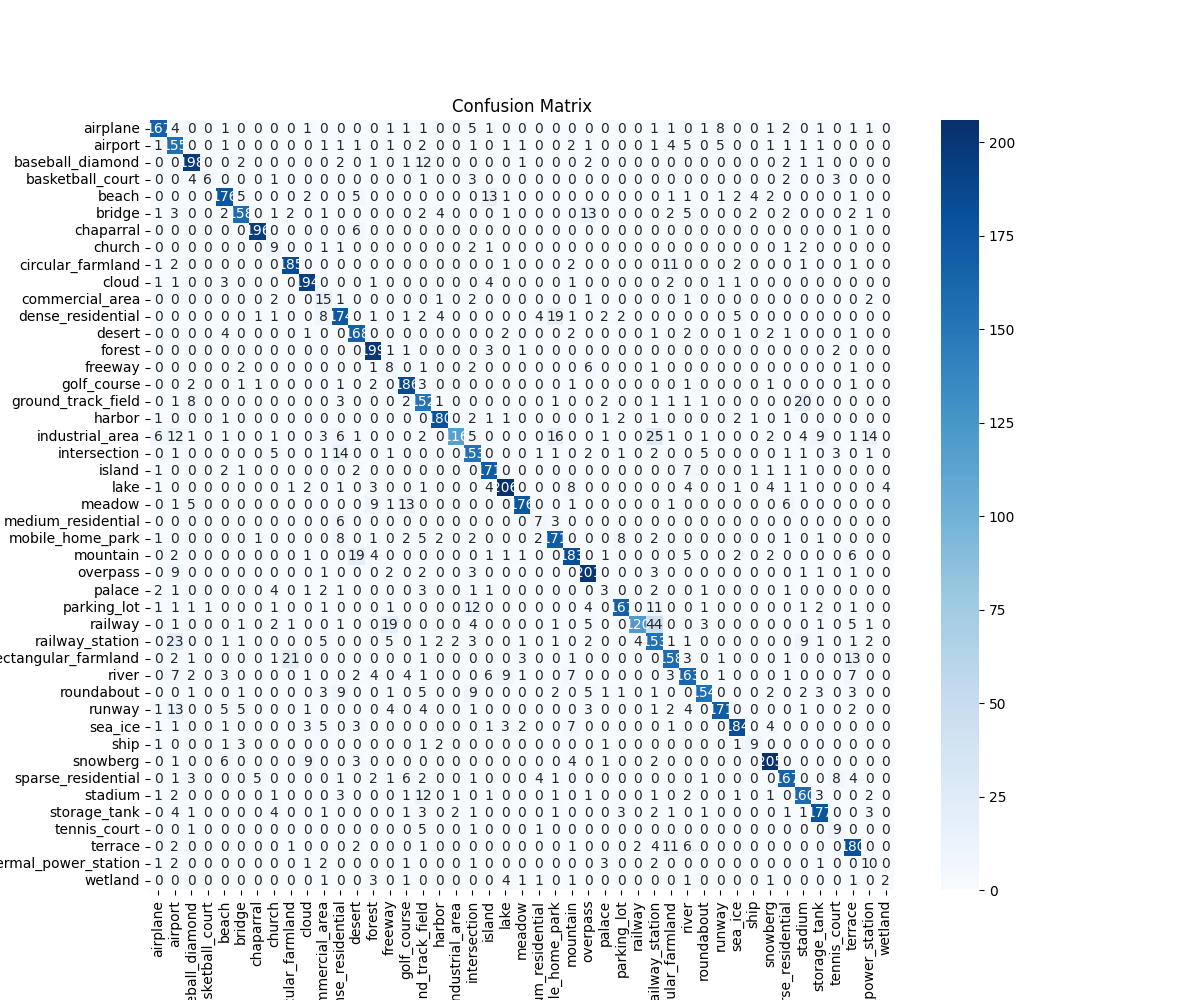}
\caption{The confusion matrix results of PatternNet under the FedAvg method.}
\label{confusion2}
\end{minipage}%
}%
\centering
\end{figure*}

The confusion matrices in Fig. \ref{confusion1} and Fig. \ref{confusion2} illustrate the classification performance of SAFE compared to the FedAvg baseline on the NWPU-RESISC45 and PatternNet datasets. The results demonstrate that SAFE consistently achieves optimal classification outcomes, with minimal misclassifications across all categories, indicating its effectiveness in addressing classification accuracy challenges for both simple and complex classes. In contrast, the baseline FedAvg method exhibits frequent misclassifications for certain categories, such as railway and industrial areas, often confusing them with visually similar classes like highways or railway stations, revealing its limitations in handling challenging samples. SAFE addresses these limitations through its integrated mechanisms, including hard sample augmentation, personalized feature preservation, and foreground enhancement, enabling efficient collaborative perception of difficult samples.

\subsubsection{Training Process}

\begin{figure*}[htbp]
\centering

\subfigure{
\begin{minipage}[t]{0.5\linewidth}
\centering
\includegraphics[width=3.7in]{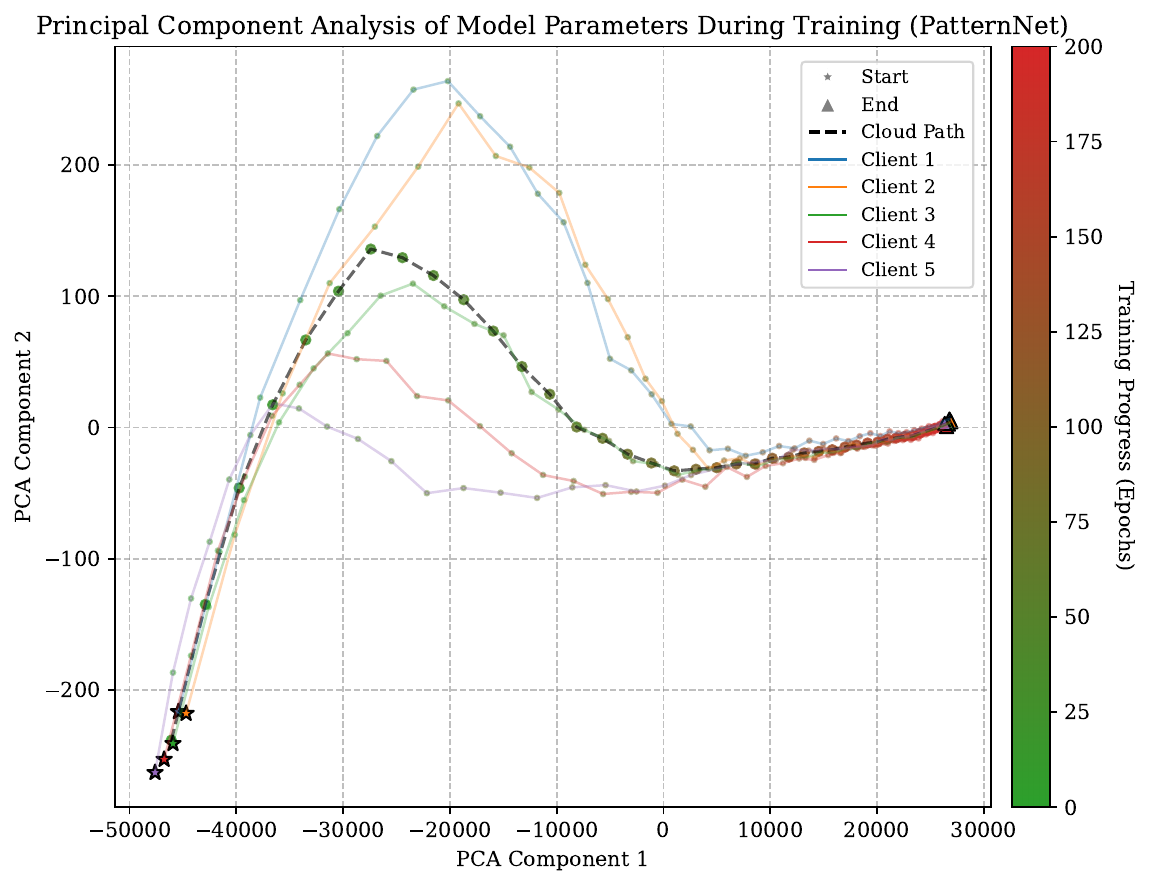}
\caption{Principal Component Analysis (PCA) was employed to extract the first two principal components of the models from five clients and the cloud, visualizing the training process of the SAFE framework on the PatternNet classification dataset.}
\label{traj1}
\end{minipage}%
}%
\subfigure{
\begin{minipage}[t]{0.5\linewidth}
\centering
\includegraphics[width=3.7in]{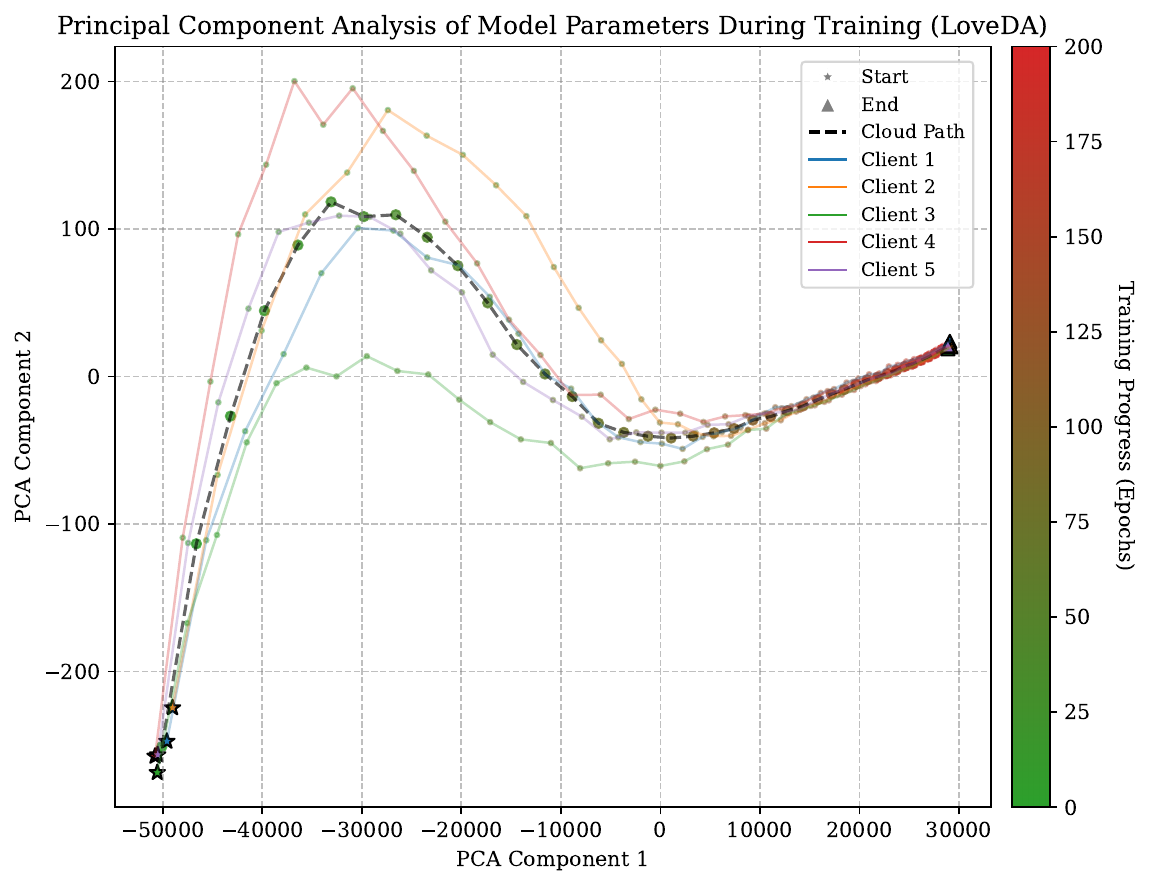}
\caption{Principal Component Analysis (PCA) was employed to extract the first two principal components of the models from five clients and the cloud, visualizing the training process of the SAFE framework on the LoveDA segmentation dataset.}
\label{traj2}
\end{minipage}%
}%
\centering

\end{figure*}

To demonstrate the variation in parameter distances among clients during training and to visually illustrate the impact of the Dual-Factor Modulation Rheostat mechanism, we conducted Principal Component Analysis (PCA) on the stacked parameters of all network modules, visualizing the first two principal components, as shown in Fig.~\ref{traj1} and Fig.\ref{traj2}.

The figure presents the trajectories of parameter principal components for five client networks and the server network throughout the training process. Initially, all clients download identical model parameters from the cloud, resulting in a unified starting point. As training progresses, client parameters initially diverge due to the Feature Adaptation Unit (FAU) mechanism, which prioritizes preserving individual parameter characteristics to optimize local training performance. Subsequently, network parameters gradually converge toward the server parameters. This convergence reflects the increasing influence of class ratio correction in later training stages, leading to similar gradient directions, while the diminishing effect of the FAU mechanism promotes parameter alignment with the server model. Consequently, all endpoints achieve parameter consistency at the conclusion of training.

\subsubsection{Ratio Similarity}
\begin{figure}[h]
    \centering
    \includegraphics[width=0.95\linewidth]{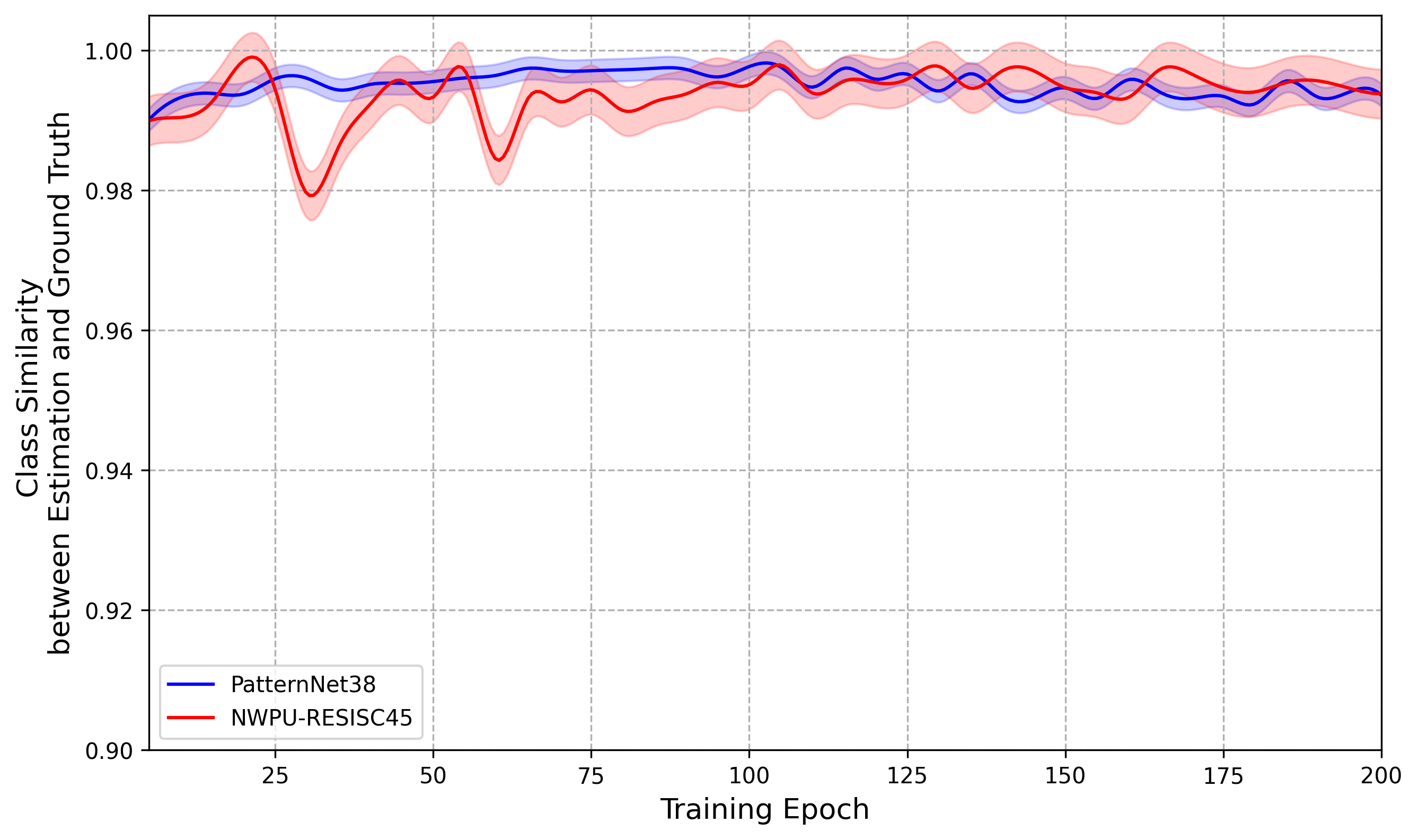}
    \caption{Ratio similarity analysis across different iterations. The visualization highlights the relationship between key parameters and their relative importance.}
    \label{fig:ratio_similarity}
\end{figure}

Fig.\ref{fig:ratio_similarity} presents the class ratios calculated based on gradient proportions in the Class Ratio Optimization (CRO) mechanism. We evaluate the cosine distance between these calculated ratios and the true label ratios across four datasets, visualizing the results throughout the training process. The analysis reveals an initial discrepancy between estimated and true class ratios during the early training stages. As training progresses, the CRO mechanism demonstrates an oscillatory convergence toward the true class ratios, indicating its capability to progressively refine the approximation to actual class distributions through gradient updates.

%
%

\section{Conclusion}
To address the challenges posed by the complex distribution of image samples from distributed satellites, such as foreground-background differences and non-i.i.d. class distributions, we propose three key strategies: (1) A Class Rectification Optimization strategy to mitigate global class imbalance by dynamically adjusting loss weights through gradient measurement without requiring clients to upload private data. (2) A Feature Alignment Update strategy to handle non-i.i.d. class distributions by updating on-board model parameters using an Exponential Moving Average (EMA) approach, preserving personalized client information within a dual-factor curriculum learning framework. (3) An Adaptive Enhanced Embedding mechanism to address intra-class variation, scale differences, and foreground-background imbalance in high-resolution remote sensing images, enhancing cross-platform collaborative feature extraction.

\section*{Acknowledgments}



 
%
\bibliography{tgrs}
\bibliographystyle{IEEEtran}

\newpage

\vspace{11pt}

\vfill

\end{document}